\documentclass[lettersize,journal]{IEEEtran}
\usepackage{amsmath,amsfonts}
\usepackage{graphicx}
\usepackage{float}
\usepackage{algorithm}
\usepackage{array}
\usepackage{textcomp}
\usepackage{stfloats}
\usepackage{url}
\usepackage{verbatim}
\usepackage{graphicx}
\usepackage{cite}
\usepackage{amsmath, amssymb, epsfig, graphicx, bm}
\usepackage{epstopdf}
\usepackage{cases}
\usepackage{float}
\usepackage[justification=centering]{caption}
\usepackage{graphicx}
\usepackage{color}
\usepackage{epstopdf}
\usepackage{amsthm}
\usepackage{amsfonts}
\usepackage{cite, url}
\usepackage[all,cmtip]{xy}
\usepackage{color,hyperref}
\usepackage[noend]{algpseudocode}
\usepackage{flushend}
\usepackage[normalem]{ulem} 

\newtheorem{theorem}{Theorem}
\newcommand{\E}{{\mathbb{E}}}
\newcommand{\R}{{\mathbb{R}}}
\newcommand{\RR}{{\mathcal{R}}}
\newcommand{\B}{{\mathcal{B}}}

\newtheorem{Definition}{Definition}

\newtheorem{assumption}{Assumption}
\newtheorem{lemma}{Lemma}
\newtheorem{remark}{Remark}
\newtheorem{corollary}{Corollary}

\newcommand{\vx}{\bm{x}}

\newcommand{\vz}{\bm{z}}
\newcommand{\ve}{\bm{e}}

\newcommand{\N}{\mathcal{N}}

\newcommand{\cM}{\mathcal{M}}
\newcommand{\cH}{\mathcal{H}}

\newcommand{\cS}{\mathcal{S}}

\newcommand{\A}{\mathcal{A}}

\newcommand{\lp}{\left(}
\newcommand{\rp}{\right)}

\newcommand{\lnorm}{\left\|}
\newcommand{\rnorm}{\right\|}

\newcommand{\smallCE}[1]{$<$1e-07}

\begin{document}
\title{On the Tradeoff between Privacy Preservation and Byzantine-Robustness in Decentralized Learning}

\author{Haoxiang Ye, Heng Zhu, and Qing Ling
\thanks{Haoxiang Ye and Qing Ling are with the School of Computer Science and Engineering, Sun Yat-Sen University, Guangzhou, Guangdong, China 510006.
        (e-mail: yehx9@mail2.sysu.edu.cn; lingqing556@mail.sysu.edu.cn)}%
\thanks{Heng Zhu is with the Department of Electrical and Computer Engineering, University of California, San Diego, La Jolla, California, USA 92093.
        (e-mail: hez007@ucsd.edu)}
\thanks{
Qing Ling (corresponding author) is supported in part by NSF China grant 62373388, Guangdong Basic and Applied Basic Research Foundation grants 2021B1515020094 and 2023B1515040025, and R\&D project of Pazhou Lab (Huangpu) grant 2023K0606.
A short, preliminary version of this paper has been presented in International
Conference on Acoustics, Speech and Signal Processing, 2024 \cite{ye2024tradeoff}.}
}



\maketitle

\begin{abstract}
This paper jointly investigates privacy preservation and Byzantine-robustness in decentralized learning. In a decentralized network, honest-but-curious agents faithfully follow the prescribed algorithm, but expect to infer their neighbors' private data from messages received during the learning process, while dishonest-and-Byzantine agents disobey the prescribed algorithm, and deliberately disseminate wrong messages to their neighbors for the sake of biasing the learning process. For this novel setting, we investigate a generic privacy-preserving and Byzantine-robust decentralized stochastic gradient descent (SGD) framework, in which Gaussian noise is injected to preserve privacy and robust aggregation rules are adopted to counteract Byzantine attacks. We analyze its learning error and privacy guarantee, discovering an essential tradeoff between privacy preservation and Byzantine-robustness in decentralized learning -- the learning error caused by defending against Byzantine attacks is exacerbated by the Gaussian noise added to preserve privacy.
For a class of state-of-the-art robust aggregation rules, we give unified analysis of their ``mixing abilities''. Building upon this analysis, we reveal how the ``mixing abilities'' affect the tradeoff between privacy preservation and Byzantine-robustness. The theoretical results provide guidelines for achieving a favorable tradeoff with proper design of robust aggregation rules. Numerical experiments are conducted and corroborate our theoretical findings.
\end{abstract}


\section{Introduction}
\IEEEPARstart{R}{ecent} years have witnessed the dramatic increases of distributed devices and the ensuing amount of distributed data. Consequently, there is a growing need to collectively train various machine learning models from distributed data over distributed devices. While traditional federated learning coordinated with a central server is still popular, decentralized learning that uses autonomous collaboration to bypass a central server has attracted much attention from both academia and industry \cite{Lian2017,Ye2021,Ding2021,10429818}. In decentralized learning, devices (also termed as agents) only need to exchange intermediate models with their neighbors, which avoids the communication bottleneck in traditional federated learning where the central server exchanges models with all participating agents.
Nevertheless, two potential security threats still limit the implementations of decentralized learning: privacy leakage and Byzantine attacks.

In exchanging the intermediate models with their neighbors, the agents may disclose their local private data\cite{Zhu2019,10639355}. To handle this challenge, one popular approach is adding random noise to the transmitted messages for preserving privacy. The works of \cite{wang2023tailoring} and \cite{huang2019dp} respectively propose to add Gaussian noise to decentralized stochastic gradient descent (SGD) and alternating direction method of multipliers (ADMM), yielding their differentially private variants. Further modifications of differentially private decentralized ADMM are in \cite{Huang2020,8871117,9666883}.


Meanwhile, the agents participating in decentralized learning are not always honest. Some agents may deliberately transmit wrong messages to their neighbors so as to bias the learning process.  We characterize such behaviors using the Byzantine attacks model -- wrong messages from Byzantine agents can be arbitrarily malicious, and identities of Byzantine agents are unknown \cite{lamport}.
To counteract Byzantine attacks, various
robust aggregation rules have been proposed to replace the classical weighted mean aggregation rule in decentralized SGD \cite{yang2019byrdie,fang2022bridge,december,he2022byzantine,wu2022byzantine}. They are proved effective to mitigate the impact of Byzantine attacks, both theoretically and experimentally.

To the best of our knowledge, these two security threats are often treated separately within decentralized learning, and no existing work simultaneously addresses the both.
Their interplay leads to remarkable difficulties, as the introduction of randomness through privacy-preserving mechanisms complicates the task of handling Byzantine agents.
Some recent works investigate both issues in federated learning with a central server
\cite{10093038,10473093,dong2021flod,10250809,10251424,ijcai2022p337,ma2022differentially,guerraoui2021differential,Allouah:302948,10423198}
but their algorithms and results do not apply in the decentralized setting.
Among these works, \cite{10093038,dong2021flod,10250809,10251424,10473093} are based on homomorphic encryption and secure multi-party computation, while \cite{ijcai2022p337,ma2022differentially,guerraoui2021differential,Allouah:302948,10423198} are based on differential privacy as in our work.
However, the absence of a central server brings us new challenges.
 {In federated learning the central server has a global model, whereas in decentralized learning each agent maintains a local model. For the latter, reaching consensus of the local models is nontrivial in the presence of Byzantine attacks, even without considering privacy preservation. It has been shown that some popular robust aggregation rules may lead to large consensus errors, and consequently, large learning errors \cite{wu2022byzantine}. Such learning errors are further exacerbated by the randomness introduced for privacy preservation.} These new challenges call for new understandings to the tradeoff between privacy preservation and Byzantine-robustness in decentralized learning, as well as new guidelines to design favorable algorithms.

{\color{black}Compared to the short and preliminary conference version \cite{ye2024tradeoff}, this paper has been extensively revised. We have incorporated comprehensive analysis of the ``mixing abilities''  for a class of state-of-the-art robust aggregation rules. Additionally, we have discussed related works in depth and included detailed proofs of the main theorems. Last but not the least, we have augmented this paper with extra numerical experiments, along with discussions on the results.}

\subsection{Our Contributions}

We summarize our contributions as follows.

\noindent \textbf{C1)} We jointly consider privacy preservation and Byzantine-robustness in decentralized learning, which, to our knowledge, has not been investigated before.

\noindent \textbf{C2)} We consider a generic privacy-preserving and Byzantine-robust decentralized SGD framework that uses the Gaussian noise mechanism to preserve privacy and robust aggregation rules to counteract Byzantine attacks. Our analysis reveals that an essential tradeoff exists between privacy preservation and Byzantine-robustness in decentralized learning -- no one can enjoy the best of the two.

\noindent \textbf{C3)} We analyze the ``mixing abilities'' of various robust aggregation rules and discover how they affect the tradeoff between privacy preservation and Byzantine-robustness. The analysis provides guidelines for achieving a favorable tradeoff -- the robust aggregation rules should be properly designed to have superior ``mixing abilities''.


\noindent \textbf{C4)} We conduct numerical experiments to demonstrate the performance of the proposed generic algorithmic framework. The numerical results corroborate our theoretical findings.

\subsection{Related Works}
\noindent\textbf{Privacy Preservation.} One of the primary challenges faced by decentralized learning is privacy preservation. Although decentralized learning allows all the agents to keep their private data locally, the intermediate models exchanged between neighboring agents may lead to serious privacy leakage \cite{Zhu2019}.
To address this concern, \cite{wang2023tailoring,huang2019dp,Huang2020,8871117,9666883} utilize differential privacy-based techniques to mitigate the risk of privacy leakage.
Aiming at guaranteeing exact convergence to an optimal solution, \cite{wang2023tailoring} designs a sequence of weakening factors to alleviate the negative impact of the added noise while \cite{huang2019dp} uses diminishing noise variance. In \cite{Huang2020}, updates of multiple variables in decentralized ADMM are added with calibrated noise, which helps improve the tradeoff between privacy preservation and convergence accuracy.
In \cite{8871117}, differentially private decentralized ADMM is modified by updating odd and even iterations in different ways -- the even-iteration variables are computed according to the previous odd-iteration variables, not directly from the private data. For decentralized nonsmooth learning, \cite{9666883} uses zeroth-order approximations in decentralized ADMM. The randomness from the zeroth-order approximations helps preserve data privacy.
Since additive noise generally affects convergence accuracy, some works introduce time-varying variables (namely, multiplicative noise) to conceal the agents' local private data. The works of \cite{GWang2022,DWang2022,GAO2023110878} assign the agents with time-varying mixing matrices and heterogeneous step sizes that are locally generated. In \cite{lou2017privacy}, data privacy is preserved through projection operations and asynchronous heterogeneous step sizes. Other existing approaches to privacy preservation include homomorphic encryption \cite{10506637,10334062}
and secure multi-party computation \cite{lu2022privacy,so2020scalable}.
However, these approaches often suffer from high computation and/or communication complexity.

\noindent\textbf{Byzantine-Robustness.} On the other hand, some agents can manipulate the messages to transmit (termed as Byzantine attacks) so as to bias the learning process. Although Byzantine-robustness has been intensively discussed in federated learning with a central server
\cite{yin2018byzantine,chen2017distributed,blanchard2017machine,cao2020fltrust,allouah2024tackling,9468910,10398506}, its study in decentralized learning is still lacking. The work of \cite{YAN2022110530} investigates decentralized consensus and \cite{9147396,8359183,9137648} consider decentralized deterministic optimization, both different to the decentralized stochastic optimization problem that we focus on. In the context of decentralized learning, \cite{yang2019byrdie,fang2022bridge} let each non-Byzantine agent utilize trimmed mean to aggregate received messages so as to reduce the impact of Byzantine agents. The work of \cite{december} uses the total variation norm to penalize the difference between neighboring models, allowing them to be different but limiting the influence of outliers. In self-centered clipping (SCC), each non-Byzantine agent uses its local model as the center and clips the received messages that disagree significantly \cite{he2022byzantine}. The work of \cite{wu2022byzantine} studies conditions of the robust aggregation rule, under which non-Byzantine agents can reach consensus and attain a small learning error. Iterative outlier scissor (IOS) is shown to be such a rule, where each non-Byzantine agent maintains a trusted set and iteratively discards neighboring models that deviate from the average model of the trusted set.

\subsection{Paper Organization}
The rest of this paper is organized as follows. Section \ref{s2} overviews the decentralized learning task, taking into account both privacy preservation and Byzantine-robustness. A generic privacy-preserving and Byzantine-robust decentralized SGD framework is introduced in Section \ref{s3}, followed by analysis of learning error and privacy guarantee in Section \ref{s4}. In Section \ref{s5}, we reveal the essential tradeoff between privacy preservation and Byzantine-robustness and provide guidelines for designing algorithms to achieve favorable tradeoffs. Section \ref{s6} discusses various existing robust aggregation rules and evaluates their tradeoffs when embedded into the privacy-preserving and Byzantine-robust decentralized SGD framework. Numerical experiments are presented in Section \ref{s7}, and conclusions are made in Section \ref{s8}.

\section{Problem Statement}
\label{s2}
We consider a decentralized network characterized by an undirected and connected graph $\mathcal{G}=(\mathcal{N},\mathcal{E})$, where $\mathcal{N}=\left\{1,...,N\right\}$ stands for the set of agents, and $\mathcal{E}$ stands for the set of edges. If $e=(m,n) \in \mathcal{E}$, then agents $m$ and $n$ are neighbors and can communicate with each other. The agents collaboratively fulfill a learning task that we will define later. During the learning process, there are two different security threats inside the decentralized network.

\noindent 1) \textit{Honest-but}-\textbf{curious agents}. Curious agents faithfully follow the prescribed decentralized learning algorithm, but expect to infer their neighbors' private data through gathered messages during the learning process.

\noindent 2) \textit{Dishonest-and}-\textbf{Byzantine agents}. Byzantine agents disobey the prescribed decentralized learning algorithm, and deliberately send arbitrarily wrong messages to their neighbors so as to bias the learning process.

Here we emphasize that all the agents are possibly curious, whereas some of them are Byzantine. The identities of Byzantine agents are unknown.
The curious agents are able to collude with each other, and so are the Byzantine agents. Our goal is to jointly preserve privacy and defend against Byzantine attacks for decentralized learning.
This practical and difficult setting has not been investigated in prior works, to the best of our knowledge.

Denote $\mathcal{R}$ and $\mathcal{B}$ respectively as the sets of non-Byzantine and Byzantine agents, and $\N = \RR \cup \B$.
For agent $n$, denote $\mathcal{R}_n$ and $\mathcal{B}_n$ respectively as the sets of non-Byzantine and Byzantine neighbors, and $\N_n = \RR_n \cup \B_n$.
The decentralized learning task is to minimize the average of the non-Byzantine agents' local aggregated  {costs (also known as objective functions or training losses)}, given by
\begin{equation}\label{1}
    \vx^*=\arg\min\limits_{\vx \in \R^D} \frac{1}{|\RR|}\sum\limits_{n \in \RR}\mathbb{E}_{\xi_n} f_n\left(\vx; \xi_n\right).
\end{equation}
Here $\vx$ is the model to learn and $f_n\left(\vx,\xi_n\right)$ is the local cost of non-Byzantine agent $n$ with respect to random variable $\xi_n\sim \cS_n$.  {In decentralized learning, $\cS_n$ corresponds to the distribution of non-Byzantine agent $n$'s local dataset, while $\xi_n$ corresponds to a mini-batch of data samples.}
The sizes of the local datasets are all assumed to be $S$, without loss of generality.
For notational convenience, define $f_n\left(\vx\right) := \mathbb{E}_{\xi_n} f_n\left(\vx; \xi_n\right)$ and $f\left(\vx\right) := \frac{1}{|\RR|} \sum_{n \in \RR} f_n\left(\vx\right)$ respectively as the local aggregated cost of non-Byzantine agent $n$ and the global aggregated cost. Note that we should not include Byzantine agents in \eqref{1} because they can arbitrarily manipulate their local aggregated costs.


\begin{remark}
The privacy threat is from the curious agents. Several existing works also call it \textit{internal threat} \cite{huang2019dp,GWang2022,DWang2022,lu2022privacy,so2020scalable}. On the other hand, \textit{external threat} has been investigated in \cite{wang2023tailoring,huang2019dp,GWang2022,DWang2022}. Therein, an eavesdropper observes all messages transmitted over the network and attempts to infer the private data. We allow all the agents to be curious and collude, which is equivalent to the case where an eavesdropper is present. Hence, our proposed algorithm is able to handle both internal and external privacy threats.

 {Since we focus on a class of SGD algorithms, the private information to protect are non-Byzantine agents' stochastic gradients. As such, a multitude of gradient inversion attacks \cite{Zhu2019}, which aim to infer raw data samples from the stochastic gradients, will be rendered ineffective.}
\end{remark}

{\color{black}
The major notations used throughout this paper are summarized in Table \ref{table:notation}.
\begin{table}[!htbp]\color{black}
    \caption{\color{black} Major notations used throughout this paper}
    \label{table:notation}
    \centering
    \begin{tabular}{cccc}
    \hline
    Symbol  &Meaning \\
    \hline
    $\vx$ & model to learn \\
    $\bar\vx$ & average model of non-Byzantine agents \\
    $\N$ &  set of agents  \\
    $\RR$ &  set of non-Byzantine agents  \\
    $\B$ &  set of Byzantine agents \\
    $f$   & global aggregated cost  \\
    $\alpha$   & step size  \\
    $B$   & batch size  \\
    $C$   & noise level  \\
    $D$   & dimensionality of model $\vx$  \\
    $M$   & stochastic gradient bound  \\
    $H$   & disagreement measure  \\
    $W$   & virtual mixing matrix  \\
    $\rho$ & contraction constant \\
    $\chi^2$   & skewness of $W$  \\
    $\epsilon$ & privacy loss \\
    $\delta$ & probability of failure \\
    $\mu$ & strong convexity coefficient  \\
   {\color{black} $\tau$} &  {\color{black}clipping threshold} \\
    $\bm{1}$ &  $|\mathcal{R}|$-dimensional all-one vector \\
    $\N_n$ &  set of neighbors of agent $n$ \\
    $\RR_n$ &  set of non-Byzantine neighbors of agent $n$  \\
    $\B_n$ &  set of Byzantine neighbors of agent $n$ \\
    $\vx_n$ & local model of agent $n$ \\
    $\tilde\vx_n$ & noisy local model of agent $n$ \\
    $\bm{e}_n$ & noise vector added to $\vx_n$ \\
    $\sigma_n^2$ & variance of $\bm{e}_n$ \\
    $f_n$   & local aggregated cost of agent $n$ \\
    $\mathcal{S}_n$   & distribution of local dataset of agent $n$ \\
    $\xi_n$ & random variable associated with data sampling in $\mathcal{S}_n$ \\
    $\mathcal{A}_n$ & robust aggregation rule applied to agent $n$ \\
    \hline
    \end{tabular}
\end{table}
}

\section{Privacy-preserving and Byzantine-robust Decentralized Learning}
\label{s3}
We consider a generic privacy-preserving and Byzantine-robust decentralized learning framework, as outlined in Algorithm \ref{dprobust}, to solve \eqref{1}. Each non-Byzantine agent $n \in \RR$ holds a local copy $\vx_n$ of the model $\vx$, which is initialized as $\vx_n^0$. {\color{black}At iteration $k$, it accesses a mini-batch of $B \leq S$ data samples according to a random set $\xi_n^{k}$,
averages the clipped stochastic gradients as
        $\tilde\nabla f_n(\vx^{k}_n; \xi_n^{k}) = \frac{1}{B} \sum_{\xi \in \xi_n^k } Clip(\nabla f_n(\vx^{k}_n; \xi),\tau)$ where $Clip(z,\tau) := z \cdot \min\{1,\frac{\tau}{\|z\|}\}$ and $\tau > 0 $ is a clipping threshold,
and then computes a noisy local model $\tilde{\vx}_{n,n}^{k} = \vx^k_n - \alpha^{k} \tilde\nabla f_n(\vx^{k}_n; \xi_n^{k}) + \mathbf{e}_n^{k }$ where the step size $\alpha^k>0$ and the noise vector $\mathbf{e}_n^{k }$ follows multivariate Gaussian distribution with mean 0 and variance $(\sigma^k)^2I$.} After that, it sends the noisy local model $\tilde{\vx}^{k}_{n, m}=\tilde{\vx}_{n,n}^{k}$ to each neighbor $m$. The purpose of adding noise is to preserve the privacy of its local data samples (in terms of their stochastic gradients). Meanwhile, each Byzantine agent $n$ sends $\tilde{\vx}^{k}_{n, m}=*$ to each neighbor $m$, with $*$ being an arbitrary $D$-dimensional vector. Upon receiving all messages from its neighbors, each non-Byzantine agent $n \in \RR$ updates its model as $\vx^{k+1}_n= \mathcal{A}_n (\tilde{\vx}_{n,n}^{k}, \{\tilde{\vx}_{m,n}^{k}\}_{m\in \RR_n\cup \B_n})$, in which $\mathcal{A}_n$ stands for a aggregation rule robust to Byzantine attacks, including trimmed mean \cite{fang2022bridge}, SCC \cite{he2022byzantine}, IOS \cite{wu2022byzantine}, to name a few.

Algorithm \ref{dprobust} reduces to privacy-preserving decentralized SGD when no Byzantine agents are present and the robust aggregation rules $\{\mathcal{A}_n\}_{n\in \RR}$ ensure proper weighted means. If privacy is not a concern and noise is not added, Algorithm \ref{dprobust} recovers Byzantine-robust decentralized SGD with robust aggregation rules.

\begin{algorithm}[t]
    \caption{Privacy-Preserving and Byzantine-Robust Decentralized Stochastic Gradient Descent (SGD)}
    \label{dprobust}
        \textbf{Input:} Initialized local model $\vx^0_n$ and Gaussian noise $\mathbf{e}_n^{k } \sim N(0,(\sigma^k)^2I) \in \R^D$ for each $n \in \RR$; Step size $\alpha^k$; Clipping threshold $\tau$
        \begin{algorithmic}[1]
        \ForAll {$k = 0, 1, 2, \cdots$}
        \ForAll {non-Byzantine agents $n \in \RR$}
        \State {\color{black}Sample a random set $\xi_n^k$ of size $B$ from $\cS_n$
        }
        \State {\color{black} Average clipped stochastic gradients as
        $$\hspace{0.8em} \tilde\nabla f_n(\vx^{k}_n; \xi_n^{k}) = \frac{1}{B} \sum_{\xi \in \xi_n^k } Clip(\nabla f_n(\vx^{k}_n; \xi),\tau)$$
        \hspace{2.6em} where $Clip(z,\tau) := z \cdot \min\{1,\frac{\tau}{\|z\|}\}$
        }
        \State Compute $\tilde{\vx}_{n,n}^{k} = \vx^k_n - \alpha^{k} \tilde \nabla f_n(\vx^{k}_n; \xi_n^{k})+\mathbf{e}_n^{k }$
        \State Send $\tilde{\vx}_{n,m}^{k} = \tilde{\vx}_{n,n}^{k}$ to all neighbors $m$
        \State Receive $\tilde{\vx}_{m,n}^{k}$ from all neighbors $m$
        \State Update $\vx^{k+1}_n= \mathcal{A}_n (\tilde{\vx}_{n,n}^{k}, \{\tilde{\vx}_{m,n}^{k}\}_{m\in \RR_n\cup \B_n})$
        \EndFor
        \ForAll {Byzantine agents $n\in \B$}
        \State Send $\tilde{\vx}^{k}_{n, m}=*$ to all neighbors $m$
        \EndFor
        \EndFor
        \end{algorithmic}
\end{algorithm}

Given the number of non-Byzantine agents $|\mathcal{R}|$ and the number of Byzantine agents $|\mathcal{B}|$, ideal robust aggregation rules $\{\mathcal{A}_n\}_{n\in \RR}$ should properly mix the messages from the non-Byzantine agents, as if the Byzantine agents were not present. Following \cite{wu2022byzantine}, we characterize their ``mixing abilities'' with a contraction constant and a virtual mixing matrix. Similar contraction properties can be found in \cite{he2022byzantine}, too.

\begin{Definition} (Virtual mixing matrix\footnote{For notational convenience, we assume that the non-Byzantine agents are numbered from $1$ to $|\RR|$.} and contraction constant of $\{\mathcal{A}_n\}_{n \in \RR}$).
    \label{definition:mixing-matrix}
    Consider $W \in \mathbb{R}^{\vert \RR \vert \times \vert \RR \vert}$ whose $(n,m)$-th entry $w_{nm} \in [0, 1]$ if $m \in \RR_n \cup\{n\}$ and $w_{nm} = 0$ if $m \in \RR$ but $m \notin \RR_n \cup\{n\}$, for $n \in \RR$. Further, $\sum_{m\in \RR_n \cup \{n\}}w_{nm}=1$. Define weighted average $\hat \vx_n := \sum_{m \in \RR_n\cup\{n\}}w_{nm} \tilde{\vx}_{m,n}$. If there exists a constant $\rho \geq 0$ for any $n \in \RR$ such that
    \begin{align}
        \label{inequality:robustness-of-aggregation-local}
             & \|\mathcal{A}_n (\tilde{\vx}_{n,n}, \{\tilde{\vx}_{m,n}\}_{m\in \RR_n\cup \mathcal{B}_n} )-\hat \vx_n \| \\
        \leq & \rho \max_{m \in \RR_n \cup \{n\}}\|\tilde{\vx}_{m,n} - \hat \vx_n\|, \notag
    \end{align}
    then $W$ is the virtual mixing matrix and $\rho$ is the contraction constant associated with the set of robust aggregation rules $\{\mathcal{A}_n\}_{n\in \RR}$.
\end{Definition}

When $\rho = 0$, Algorithm \ref{dprobust} performs like privacy-preserving decentralized SGD on the subgraph of non-Byzantine agents with a virtual mixing matrix $W$.
The contraction constants are usually nonzero and the virtual mixing matrices are often row stochastic but not doubly stochastic. We use $\chi^2 := \frac{1}{\vert \RR \vert}\|W^\top \bm{1}- \bm{1} \|^2$ to describe the skewness (namely, non-doubly stochasticity) of $W$, with $\bm{1}$ representing a $|\mathcal{R}|$-dimensional all-one vector.

\begin{remark}
    Definition \ref{definition:mixing-matrix} is able to cover various state-of-the-art robust aggregation rules. Remarkable examples include trimmed mean \cite{fang2022bridge}, SCC \cite{he2022byzantine} and IOS \cite{wu2022byzantine}, which are popular in decentralized learning.
    In Section \ref{s6}, we will analyze their contraction constants $\rho$ and virtual mixing matrices $W$. Note that the pair of $(\rho, W)$ are not unique and finding the best pair is out of the scope of this paper.
\end{remark}

\section{Learning Error and Privacy Guarantee}
\label{s4}
Below we analyze the learning error and privacy guarantee of Algorithm \ref{dprobust}.
All the proofs are deferred to the supplementary material of this paper.
We begin with several assumptions.

First, the subgraph consisting of all the non-Byzantine agents must be connected, which is a common assumption in Byzantine-robust decentralized algorithms \cite{8359183,9137648,yang2019byrdie,fang2022bridge,he2022byzantine,december,9147396,YAN2022110530,wu2022byzantine}. Otherwise, under Byzantine attacks, the non-Byzantine
agents within different subgraphs at most learn their ``isolated'' models and cannot reach consensus.

\begin{assumption}(Network connectivity).
    \label{assumption:connection}
    The subgraph of all the non-Byzantine agents is connected, meaning that for any pair $n, m\in \RR$, there exists at least one path between them. For the virtual mixing matrix $W$ associated with $\{\mathcal{A}_n\}_{n \in \RR}$, we assume $\lambda := 1-\|(I-\frac{1}{\vert \RR \vert}\bm{1}\bm{1}^\top)W\|^2 \in (0,1]$.
\end{assumption}

Second, the following assumptions are common in analyzing decentralized stochastic algorithms.

\begin{assumption}(Independent sampling).
	\label{assumption:indSampling}
	All stochastic gradients $\nabla f_n(\vx^{k}_n; \xi_n^{k})$ are independently sampled over iterations $k=0, 1, \ldots$ and across the non-Byzantine agents $n \in \RR$. The sampling process is uniform and without replacement.
\end{assumption}

\begin{assumption}(Lower boundedness).
	\label{assumption:lower-bound}
	The overall aggregated cost $f(\vx)$ is lower bounded by $f(\vx^*)$, which is finite.
\end{assumption}

\begin{assumption}(Strong convexity).  \label{assumption:convex}
	The cost $f_n(\vx,\xi_n)$ is $\mu$-strongly convex for each non-Byzantine agent $n\in \RR$.
\end{assumption}

\begin{assumption}(Bounded stochastic gradient).
	\label{assumption:gradients}
	The stochastic gradient of the cost $f_n(\vx_n^k,\xi_n)$ with respect to any $\vx_n^k \in \mathbb{R}^D$  is bounded by $\|\nabla f_n(\vx_n^k,\xi_n) \| \le M$ for each non-Byzantine agent $n\in \RR$ at iteration $k$.
\end{assumption}

\begin{remark}
Existing works on Byzantine-robustness often assume Lipschitz continuous stochastic gradient and bounded stochastic gradient variance \cite{wu2022byzantine,december,he2022byzantine}. We do not make these assumptions, but assume bounded stochastic gradient that is common in analyzing differential privacy. In fact, all of these three assumptions are used in \cite{Ding2021,guerraoui2021differential,Allouah:302948,Xu2021}
that consider privacy preservation. Due to the differences in the assumptions, we need to develop new tools for the analysis.

{\color{black}On the other hand, clipping is often incorporated in privacy-preserving SGD to avoid estimating the stochastic gradient bound $M$. Some existing works require the clipping threshold $\tau$ to be larger than the stochastic gradient bound $M$ \cite{Ding2021,guerraoui2021differential,Allouah:302948}, implying that clipping is not actually performed. In contrast, we allow $\tau$ to be arbitrary. This more general setting complicates the analysis, as we will see in the following.}

\end{remark}

\subsection{Disagreement Measure}

Prior to investigating the learning error of the proposed Algorithm \ref{dprobust}, we need to characterize how the non-Byzantine agents reach consensus during the learning process. Define the disagreement measure as
    \begin{align}
        \label{hk}
		H^k =  \frac{1}{\vert \RR \vert}\sum_{n\in \RR} \|\vx^{k}_{n}-\bar\vx^{k}\|^2,
	\end{align}
in which $\bar\vx^k := \frac{1}{|\RR|}\sum_{n\in \RR}\vx^k_n$ is the average of all local models of the non-Byzantine agents at iteration $k$. The following theorem bounds the disagreement measure of Algorithm \ref{dprobust}.

\begin{theorem}(Disagreement measure).
    \label{t5}
    Suppose that the robust aggregation rules $\{\A_n\}_{n\in \RR}$ in Algorithm \ref{dprobust} satisfy \eqref{inequality:robustness-of-aggregation-local}, and $\rho$ satisfies
    \begin{align}
    \label{condtion:rho}
      \rho < \rho^* := \frac{\lambda}{8\sqrt{\vert \RR \vert}}.
    \end{align}
    Set the step size $\alpha^k= \frac{8}{\mu (k+k_0)}$, where $k_0$ is sufficiently large, and set $\sigma^k =C \alpha^k$. Under Assumptions \ref{assumption:connection}--\ref{assumption:gradients}, for Algorithm \ref{dprobust} we have
    \begin{align}
        \label{inequality:H-convergence}
        \E H^k
        \le & (c_1)^k H^0 + \frac{c_2 {\color{black}\varphi^2}+c_3 C^2}{\mu^2} \frac{1}{(k+k_0)^2}.
    \end{align}
    Here {\color{black}$\varphi = \min \{M,\tau\}$,} $c_1 \in (0,1)$, $c_2 > 0$, and $c_3 > 0$ are constants, and the expectation is taken over all the random variables.
\end{theorem}

From Theorem \ref{t5}, if the robust aggregation rules $\{\A_n\}_{n\in \RR}$ are properly designed such that the contraction constant $\rho$ is sufficiently small, the disagreement measure of Algorithm \ref{dprobust} diminishes over time, given that $\alpha^k = O(\frac{1}{k})$ and $\sigma^k = O(\frac{1}{k})$. The upper bound $\rho^*$ of $\rho$ is determined by $\lambda$ that characterizes the associated virtual mixing matrix $W$ and $|\RR|$ that is the number of non-Byzantine agents. For a large $k$, the second term at the right-hand side of \eqref{inequality:H-convergence} dominates and is in the order of $O(\frac{1}{k^2})$. When the added Gaussian noise is strong such that $C$ is large, the constant of this term is large too. Note that diminishing noise variance $(\sigma^k)^2=C^2(\alpha^k)^2$ is enough to preserve privacy since the step size $\alpha^k$ is also diminishing, as we will show in Section \ref{sec:IV.C}.

\subsection{Learning Error}

Next, we bound the learning error of Algorithm \ref{dprobust}.

\begin{theorem}(Learning error).
	\label{t4}
Suppose that the robust aggregation rules $\{\A_n\}_{n\in \RR}$ in Algorithm \ref{dprobust} satisfy \eqref{inequality:robustness-of-aggregation-local}. Set the step size $\alpha^k= \frac{8}{\mu (k+k_0)}$, where $k_0$ is sufficiently large, and set $\sigma^k =C \alpha^k$. Under Assumptions \ref{assumption:connection}--\ref{assumption:gradients}, for Algorithm \ref{dprobust} we have
 \begin{align}
                \label{inequality:convergence-0}
           & \hspace{1.5em} \E [f_{best}^k-f\lp\vx^*\rp]
            \leq \frac{ c_4 \mu \Vert \bar\vx^0 - \vx^*\Vert^2 }{k+1}  \\
            &+ \frac{8 ({\color{black}\varphi^2}+ DC^2/|\RR|) ln(k+k_0)}{\mu(k+1)}
             + \frac{c_5(4 \rho^2 + \chi^2)  {\color{black}\varphi^2} }{\mu}         \nonumber\\
             &+  \frac{c_6(8 ln|\RR| \rho^2 + \chi^2) DC^2}{\mu}
            + \frac{\sum_{k'=0}^{k} 4M \E \sqrt{H^{k'}}}{k+1}  \nonumber\\
            & + \frac{\sum_{k^\prime=0}^k c_7  (4|\RR|\rho^2+\chi^2)  \mu (k^\prime+k_0)^2 \E H^{k^\prime}}{k+1} \nonumber\\
            &{\color{black}+ \frac{16 \max\{M-\tau,0\}^2}{\mu }.}\notag
	\end{align}
Here  {\color{black}$\varphi = \min \{M,\tau\}$,} $f_{best}^k := \min_{k^\prime \in \{1,2,\cdots,k \}} f(\bar\vx^{k^\prime})$, $c_4>0$, $c_5>0$, $c_6 > 0$, and $c_7>0$ are constants, and the expectation is taken over all the random variables.
\end{theorem}

Theorem \ref{t4} demonstrates the convergence of the average model of all non-Byzantine agents. The first term at the right-hand side of \eqref{inequality:convergence-0} is in the order of  $O(\frac{1}{k})$, which vanishes quickly.  {The second term is in the order of $O(\frac{lnk}{k})$, and also vanishes as $k$ increases.} The third and fourth terms are related to $ \rho^2 + \chi^2$ that characterizes the impact of the contraction constant $\rho$ and the virtual mixing matrix $W$, and is also determined by the
{\color{black}clipping threshold $\tau$}
and the noise level $C$.  The fifth and sixth terms are related to the disagreement measure. As we have demonstrated in Theorem \ref{t5}, if the contraction constant $\rho$ is sufficiently small, the disagreement measure $\E H^k$ vanishes at the rate of $O(\frac{1}{k^2})$.  {In this case, the fifth term is in the order of $O(\frac{lnk}{k})$} and the sixth term is in the order of {\color{black}$O\left((\rho^2 + \chi^2) (\varphi^2 + C^2)\right)$. The seventh term comes from the bias caused by clipping, which is independent on the robust aggregation rules and the Gaussian mechanism. When the clipping threshold $\tau$ is larger than the stochastic gradient bound $M$ such that clipping is not performed, we have $\max\{M-\tau,0\}^2=0$ and the seventh term disappears. Conversely, when the clipping threshold $\tau$ is smaller than the stochastic gradient bound $M$, clipping will introduce an extra non-vanishing learning error term in the order of $O((M-\tau)^2)$, which is tight and matches the lower bound established for clipped SGD \cite{koloskova2023revisiting,li2024clipped}.} Summing up these terms, we know that the learning error $\E [f_{best}^k-f\lp\vx^*\rp]$ is in the order of {\color{black}$O\left((\rho^2 + \chi^2) (\varphi^2 + C^2) + \max\{M-\tau,0\}^2\right)$.} This result is summarized in the following corollary.

\begin{corollary}(Learning error).
	\label{coro-1}
Suppose that the robust aggregation rules $\{\A_n\}_{n\in \RR}$ in Algorithm \ref{dprobust} satisfy \eqref{inequality:robustness-of-aggregation-local} and the associated contraction constant $\rho$ satisfies \eqref{condtion:rho}. Set the step size $\alpha^k= \frac{8}{\mu (k+k_0)}$, where $k_0$ is sufficiently large, and set $\sigma^k =C \alpha^k$. Under Assumptions \ref{assumption:connection}--\ref{assumption:gradients}, for Algorithm \ref{dprobust} we have
\vspace{-1em}
	\begin{align}
                \label{inequality:convergence-1}
       & \text{Learning error} \\
       =&  {\color{black}O\left((\rho^2 + \chi^2) (\varphi^2 + C^2) + \max\{M-\tau,0\}^2\right).} \nonumber
	\end{align}
\end{corollary}

Note that when there is no Byzantine agent, the robust aggregation rules are proper such that $\rho=0$ and $\chi^2=0$, {\color{black}as well as the clipping threshold satisfies $\tau \geq M$}, Algorithm \ref{dprobust} converges precisely to the optimal solution.
{\color{black}The convergence rate is $O(\frac{lnk}{k})$, which matches the convergence rate of the classical decentralized SGD under strong convexity and bounded stochastic gradient assumptions \cite{tsianos2012distributed,nedic2016stochastic}. Actually, such a convergence rate is optimal for SGD without assuming smoothness \cite{rakhlin2011making}.}

\begin{remark}
    Our results align with the existing analysis on Byzantine-robust decentralized SGD \cite{yang2019byrdie,9137648,fang2022bridge,december,he2022byzantine,9147396,YAN2022110530,wu2022byzantine,8359183}.
     Specifically, in the presence of Byzantine agents, achieving exact convergence to the optimal solution $\vx^*$ becomes impossible. Instead, the algorithm converges to a neighborhood of $\vx^*$, and the size of this neighborhood is influenced by factors such as the topology constructed by non-Byzantine agents and the chosen robust aggregation rule. Our results shed light on how these factors explicitly influence the learning error. Therein, $\rho$ and $\chi^2$ are determined by the topology constructed by non-Byzantine agents and the chosen robust aggregation rule.

\end{remark}

\subsection{Privacy Guarantee}
\label{sec:IV.C}

To establish the privacy guarantee, we introduce two definitions about differential privacy.
\begin{Definition}[$(\varepsilon,\delta)$-differential privacy\cite{Dwork2006}]
\label{dp}
        Let $\varepsilon \geq 0$ and $\delta \in [0, 1]$.
A randomized algorithm $\mathcal{M}:\mathcal{Z}^D \rightarrow \mathcal{Y}$ satisfies $(\varepsilon, \delta)$-differential privacy if for any  adjacent inputs
$\vz, \vz' \in \mathcal{Z}^D$ and
subset $\mathcal{U} \subseteq \mathcal{Y}$, we have
\begin{equation}
    \label{eq:dp-def}
    \mathbb{P}[\mathcal{M}(\vz) \in \mathcal{U}] \leq e^{\varepsilon}\cdot \mathbb{P} \left[ \mathcal{M}(\vz') \in \mathcal{U} \right] + \delta.
\end{equation}
\end{Definition}


In the definition, the parameter $\epsilon$ represents privacy loss, which quantifies the extent of privacy leakage resulting from the algorithm. Additionally, $\delta$ denotes the probability of failure for achieving $\epsilon$-differential privacy. It is important to note that smaller values of $\epsilon$ and $\delta$ indicate stronger privacy protection offered by the algorithm. {We aim to achieve sample-level differential privacy to protect the privacy of individual data samples (in terms of their stochastic gradients) on all non-Byzantine agents.
Let $\mathcal{S}_n$ be the local dataset of non-Byzantine agent $n$. Then, $\mathcal{S}_n$ and $\mathcal{S}_n'$ are adjacent if they differ by one data sample. In decentralized learning, each non-Byzantine agent $n \in \RR$ runs a randomized algorithm $\mathcal{M}_n(\mathcal{S}_n)$, whose input is its local dataset $\mathcal{S}_n$ and output contains all transmitted messages. Our goal is to ensure that all $\mathcal{M}_n(\mathcal{S}_n)$ simultaneously satisfy $(\epsilon,\delta)$-differential privacy for all non-Byzantine agents $n \in \RR$. A general idea to achieve sample-level differential privacy is to add Gaussian noise to transmitted messages in the decentralized learning process.}


In the analysis, we leverage the concept of R{\'e}nyi differential privacy \cite{mironov2017renyi}.

\begin{Definition}(R{\'e}nyi differential privacy).
    A randomized algorithm $\cM$
    is $r$-R{\'e}nyi differentially private of order $g$, if for any adjacent inputs $\vz,\vz' $ it holds that
    \begin{equation*}
        D_g\left(\cM(\vz) || \cM(\vz') \right) \leq r,
    \end{equation*}
    where $D_g$ denotes the R{\'e}nyi divergence of order $g > 1$.
\end{Definition}

Now, we give the privacy guarantee of Algorithm \ref{dprobust}.
{\color{black}
First, if the batch size $B$ is the same as the sample size $S$, then the Gaussian noise mechanism $\tilde{\vx}_{n,n}^{k} = \vx^k_n - \alpha^{k} \tilde \nabla f_n(\vx^{k}_n; \xi_n^{k}) + \mathbf{e}_n^{k }$, where $\mathbf{e}_n^{k} \sim N(0,(C\alpha^k)^2I)$, is guaranteed to be $(g,\frac{2g  \varphi^2}{C^2 B^2})$-R{\'e}nyi differentially private of any order $g>1$ at each iteration $k$, according to \cite{mironov2017renyi}.

Second, for the random sampling scenario $B < S$, the following Lemma \ref{lemma-subsample} shows that the Gaussian noise mechanism is {$(g,\frac{14  g  \varphi^2}{C^2 S^2})$-R{\'e}nyi differentially private at each iteration $k$, when $\frac{C^2B^2}{\varphi^2} \geq 2.8$ and $g \leq - \frac{C^2 B^2}{6 \varphi^2}  \log \lp \frac{B}{S}\lp1+\frac{C^2B^2}{4\varphi^2}\rp g\rp+1$.} Although the stochastic gradient noise amplifies the privacy protection ability, it is not sufficient on its own. Extra Gaussian noise needs to be added to provide the privacy guarantee.

\begin{lemma}(Gaussian noise mechanism with random sampling \cite{wang2019subsampled,wang2023efficient}).
\label{lemma-subsample}
    Consider a Gaussian noise mechanism $\cM$ that is $(g,\frac{g}{2 u^2})$-R{\'e}nyi differentially private without random sampling. If applied to a subset of samples at the sampling rate $\frac{B}{S}$ using uniform sampling without replacement, then {it is $(g,\frac{3.5g B^2}{S^2 u^2})$-R{\'e}nyi differentially private, when $u^2 \geq 0.7$ and $g \leq - \frac{2}{3} u^2 \log (\frac{B}{S}\lp1+u^2\rp g)+1$.}
\end{lemma}

Third, according to the linear composition property \cite{mironov2017renyi}, Algorithm \ref{dprobust} is { $(g,\frac{14  g  \varphi^2 K}{C^2 S^2})$}-R{\'e}nyi differentially private given the total number of iterations $K$. Finally, we use the following Lemma \ref{lemma-convert} to convert R{\'e}nyi differential privacy into differential privacy. Setting {$g = \frac{CS}{\varphi } \sqrt{\frac{\log(1/\delta)}{14K}} +1$} yields Theorem \ref{th1}.}

\begin{lemma}(Conversion from R{\'e}nyi differential privacy to differential privacy \cite{mironov2017renyi}).
\label{lemma-convert}
    If a randomized mechanism $\cM$ is $(g,r)$-R{\'e}nyi differentially private, then it is $(r +\frac{\log\frac{1}{\delta}}{g-1},\delta)$-differentially private for any $\delta \in (0,1)$.
\end{lemma}

\begin{theorem}(Privacy guarantee).
    \label{th1}
    For Algorithm \ref{dprobust}, set the step size $\alpha^k= \frac{8}{\mu (k+k_0)}$, where $k_0$ is sufficiently large, and set $\sigma^k =C \alpha^k$.
    Under Assumption \ref{assumption:gradients},
    Algorithm \ref{dprobust} is {\color{black} $(\frac{14  \varphi^2 K}{C^2 S^2}$ $+ \frac{2\varphi}{CS} \sqrt{14K\log(\frac{1}{\delta})},\delta)$-differentially private for any failure probability $\delta \in (0,1)$ and given number of iterations $K$, when $\frac{C^2B^2}{\varphi^2} \geq 2.8$ and $g \leq - \frac{2C^2}{3}  \log (\frac{B}{S}\lp1+\frac{C^2B^2}{4\varphi^2}\rp g)+1$.}
\end{theorem}

When $K$ is sufficiently large, the established privacy loss $\frac{14  \varphi^2 K}{C^2 S^2} + \frac{2\varphi}{CS} \sqrt{14K\log(\frac{1}{\delta})}$ is in the order of
\begin{equation}
\label{eq:privacyloss}
{\color{black}\textit{Privacy~loss} = O \left( \frac{\varphi^2 K}{C^2} \right).}
\end{equation}



\vspace{-1em}

\section{Tradeoff between Privacy Preservation and Byzantine-Robustness}
\label{s5}

For privacy-preserving and Byzantine-robust decentralized SGD, {\color{black}if $\tau \geq M$,} Corollary \ref{coro-1} establishes the learning error in the order of {\color{black}$O\left((\rho^2+\chi^2)(\varphi^2 + C^2)\right)$}, which is a typical measure of Byzantine-robustness. On the other hand, Theorem \ref{th1} shows the privacy loss in the order of $O\left(\frac{\varphi^2K}{C^2} \right)$. {\color{black}If $\tau < M$, there will be an extra term in the learning error, determined by clipping and irrelevant with the robust aggregation rules and the Gaussian mechanism.} These theoretical results show an essential tradeoff between privacy preservation and Byzantine-robustness, and provide guidelines for designing algorithms to achieve a favorable tradeoff.

First, given any robust aggregation rule, to achieve a small privacy loss, we must set a small total number of iterations $K$ and a large noise level $C$. This enlarges the learning error and reduces Byzantine-robustness. On the other hand, to achieve a small learning error, $K$ should be sufficiently large and $C$ should be small, yielding poor privacy preservation performance.
These conclusions align with the observation in Byzantine attack-free differential privacy, namely, employing a high level of noise will hurt the utility.

Second, different to the observation in Byzantine attack-free differential privacy, our analysis further highlights how the robust aggregation rule impacts the tradeoff between privacy preservation and Byzantine-robustness.
The effect of a given noise level $C$ on the learning error is amplified by $\rho^2+\chi^2$, which varies across different robust aggregation rules.
Therefore, smaller $\rho$ and $\chi^2$ offer a better tradeoff between privacy preservation and Byzantine-robustness.
In other words, a robust aggregation rule with a superior ``mixing ability''  helps mitigate the influence of the additive Gaussian noise and leads to a favorable tradeoff.
This fact shows the importance of designing a proper robust aggregation rule that has a small contraction constant $\rho$ and a low skewness $\chi^2$ of the associated virtual mixing matrix $W$.

 {
\begin{remark} The works of \cite{nguyen2022flame,sun2019can} consider targeted attacks, which poison some data samples so that the trained model performs poorly on targeted inputs, but normally on others. Adding noise ensures that the trained model is less impacted by a small amount of poisoned data samples, and thus benefits robustness to targeted attacks. In contrast, our work, as well as \cite{yang2019byrdie,fang2022bridge,december,he2022byzantine,wu2022byzantine} and \cite{yin2018byzantine,chen2017distributed,blanchard2017machine}, focus on the worst-case Byzantine attacks. Byzantine agents can send arbitrarily malicious messages at every iteration of the learning process, not only poison their data samples in the beginning. The randomness of added noise makes it more difficult to distinguish Byzantine messages from true messages, complicating the task of achieving Byzantine-robustness.

\end{remark}
}

\section{Contraction Constants and Virtual Mixing Matrices of Robust Aggregation Rules}
\label{s6}

In Section \ref{s5}, we have already highlighted the importance of designing a robust aggregation rule with small $\rho$ and $\chi^2$ to the tradeoff between privacy preservation and Byzantine-robustness. In this section, we analyze the contraction constants and virtual mixing matrices for various state-of-the-art robust aggregation rules.


\subsection{Robust Aggregation Rules}

Below, we briefly review three popular robust aggregation rules in decentralized learning, including trimmed mean \cite{fang2022bridge}, SCC \cite{he2022byzantine} and IOS \cite{wu2022byzantine}. A number of robust aggregation rules that are successful in distributed learning, such as coordinate-wise median \cite{yin2018byzantine}, geometric median \cite{chen2017distributed} and Krum \cite{blanchard2017machine}, can be extended to the decentralized scenario too.  {However, their contraction constants are larger than the threshold given by \eqref{condtion:rho}, such that consensus is no longer guaranteed and the learning errors are remarkable.} We omit these extensions due to the page limit.
Prior to the review, recall that at each iteration of the privacy-preserving and Byzantine-robust decentralized SGD, each non-Byzantine agent $n \in \RR$ receives messages (trustful-but-noisy models or malicious models) from its neighbors $\RR_n\cup\B_n$. Non-Byzantine agent $n$ needs to aggregate these messages and its own noisy local model to yield a robust estimate of the true model.

\textbf{Trimmed mean} operates in a coordinate-wise manner. For each dimension $d$, each non-Byzantine agent $n$ discards $q_n$ largest and $q_n$ smallest elements of the received messages (not including its own noisy local model) before taking average, where $q_n$ estimates the number of its Byzantine neighbors. Such an operation can be written as
\begin{align}
    \label{tm}
    & [TM(\tilde{\vx}_{n,n}, \{\tilde{\vx}_{m,n}\}_{m\in \mathcal{R}_n\cup\B_n})]_d
    \nonumber \\
    =&\frac{1}{ \vert \mathcal{R}_n\cup\B_n \vert - 2q_n+1 }\sum_{m \in [\mathcal{U}_n]_d} [\tilde{\vx}_{m,n}]_d,
\end{align}
where $[\tilde{\vx}_{m,n}]_d$ represents the $d$-th element of $\tilde{\vx}_{m,n}$ and $[\mathcal{U}_n]_d$ is a subset of $\mathcal{R}_n\cup\B_n \cup \{n\}$ whose elements remain after removing the $q_n$ largest and $q_n$ smallest for dimension $d$.

For each non-Byzantine agent $n \in \RR$, \textbf{SCC} clips the received messages using its own noisy local model as the center. Given a clipping bound $\tau_n \geq 0$, if $\| \tilde{\vx}_{m,n}-\tilde{\vx}_{n,n} \| \leq \tau_n$, then $\tilde{\vx}_{m,n}$ is deemed trustful and no clipping happens. Otherwise, if $\| \tilde{\vx}_{m,n}-\tilde{\vx}_{n,n} \| > \tau_n$, then the message $\tilde{\vx}_{m,n}$ received from agent $m$ is faraway from $\tilde{\vx}_{n,n}$, and thus clipping is performed to limit the influence of $\tilde{\vx}_{m,n}$. Such an operation can be written as
\begin{align}
    \label{scc}
    & SCC(\tilde{\vx}_{n,n}, \{\tilde{\vx}_{m,n}\}_{m\in \mathcal{R}_n\cup\B_n}) \\
    =&\sum_{m\in \RR_n\cup\B_n\cup\{n\}} w_{nm}'\lp \tilde{\vx}_{n,n} + Clip(\tilde{\vx}_{m,n}-\tilde{\vx}_{n,n},\tau_n) \rp, \nonumber
\end{align}
where $Clip(z,\tau_n)=\min\{  1,\frac{\tau_n}{\Vert z \Vert} \} \cdot z$ and $w'_{nm}$ is the weight assigned by non-Byzantine agent $n$ to $m$.

For each non-Byzantine agent $n \in \RR$, \textbf{IOS} maintains a trusted set of neighbors and successively discards $q_n$ messages from the neighbors that are deemed to be malicious. The initial trusted set is defined as $\mathcal{U}_n^{(0)}=\RR_n\cup\B_n\cup\{n\}$ for each iteration.
In each inner iteration $i$, non-Byzantine agent $n$ discards the message that is the farthest away from the average message of the trusted set $\mathcal{U}^{(i)}_n$, and hence forms a new trusted set $\mathcal{U}^{(i+1)}_n$. It is worth noting that $n$ is always within $\mathcal{U}^{(i+1)}_n$ since its own local model is always trusted. After $q_n$ inner iterations, non-Byzantine agent $n$ has a trusted set $\mathcal{U}^{(q_n)}_n$. Finally, the weighted average message of the trusted set $\mathcal{U}^{(q_n)}_n$ is output as the result, given by
\begin{align}
    \label{ios}
    & IOS(\tilde{\vx}_{n,n}, \{\tilde{\vx}_{m,n}\}_{m\in \mathcal{R}_n\cup\B_n})
    \nonumber \\
    =& \frac{1}{\sum_{m\in \mathcal{U}^{(q_n)}_n} w_{nm}'} \sum_{m\in \mathcal{U}^{(q_n)}_n } w_{nm}'\tilde{\vx}_{m,n}.
\end{align}

\subsection{Contraction Constants and Virtual Mixing Matrices}

Now, we give the contraction constants and virtual mixing matrices of the above-mentioned robust aggregation rules. All the proofs are deferred to the supplementary material.

\begin{lemma}[Trimmed mean \cite{fang2022bridge}]
    \label{tm-rho}
For any non-Byzantine agent $n \in \RR$, suppose $|\mathcal{N}_n|\geq2|\mathcal{B}_n|$ and $q_n$ is set as $\vert \B_n \vert$. Then, the associated virtual mixing matrix $W$ is row stochastic and its $(n,m)$-th entry is given by
    \begin{align*}
        w_{nm} = \frac{1}{\vert \RR_n \vert+1}.
    \end{align*}
    Further, if all Byzantine messages are removed, the contraction constant is bounded by
    \begin{align}
        \label{equality:rho-of-tm-1}
        \rho \le & \max_{n\in\RR} \frac{2q_n}{\vert \N_n \vert - q_n+1}
        \min\{D^{\frac{1}{2}},(\vert \RR_n \vert+1) ^{\frac{1}{2}}\}.
    \end{align}
    Otherwise, the contraction constant is bounded by
    \begin{align}
        \label{equality:rho-of-tm-2}
        \rho \le & \max_{n\in\RR} \lp \frac{2q_n}{\vert \N_n \vert - 2q_n+1} +\frac{4q_n}{\vert \N_n \vert - q_n+1} \rp   \\
        & \cdot \min\{D^{\frac{1}{2}},(\vert \RR_n \vert+1) ^{\frac{1}{2}}\}. \nonumber
    \end{align}
\end{lemma}

\begin{lemma}[SCC \cite{he2022byzantine}]
\label{scc-rho}
    For any non-Byzantine agent $n \in \RR$, suppose $\tau_n $ is set as $$\tau_n  = \lp \frac{1}{\sum_{m \in \mathcal{B}_n} w'_{nm}} \sum_{m \in \mathcal{R}_n} w'_{nm} \lnorm  \tilde\vx_{n,n}  -  \tilde\vx_{m,n}  \rnorm ^2 \rp^{\frac{1}{2}}.$$ Then, the associated virtual mixing matrix $W$ is doubly stochastic and its $(n,m)$-th entry is given by
    \begin{align*}
        w_{nm} = \begin{cases}
            w_{nn}'+\sum_{b\in\B_n} w_{nb}', &m=n,\\
            w_{nm}', &m\neq n,\\
        \end{cases}
    \end{align*}
    and the contraction constant is bounded by
    \begin{align}
        \label{equality:rho-of-scc}
        \rho \le 4 \max_{n\in\RR}  \sqrt{ \sum_{m \in \mathcal{B}_n}w'_{nm} \sum_{m \in \mathcal{R}_n} w'_{nm}}.
    \end{align}
\end{lemma}

\begin{lemma}[IOS \cite{wu2022byzantine}]
\label{ios-rho}
For any non-Byzantine agent $n \in \RR$, suppose $q_n$ is set as $\vert \B_n \vert$ and define a set that includes the neighbors with the largest $q_n$ weights, as $$\cH_n := \underset{\cH': \cH' \subseteq \N_n \atop |\cH'|=q_n}{\arg\max} \sum_{m \in \cH'}w_{nm}'.$$ When $\sum_{m \in \cH_n}w'_{nm}< \frac{1}{3}$, the associated virtual mixing matrix $W$ is doubly stochastic and its $(n,m)$-th entry is given by
    \begin{align*}
        w_{nm} = \begin{cases}
            w_{nn}'+\sum_{b\in\B_n} w_{nb}', &m=n,\\
            w_{nm}', &m\neq n.\\
        \end{cases}
    \end{align*}
    Further, if all Byzantine messages are removed, the contraction constant is bounded by
    \begin{align}
        \label{equality:rho-of-ios-1}
        \rho \le \max_{n\in\RR}\frac{\sum_{m \in \cH_n}w'_{nm}}{1-\sum_{m \in \cH_n}w'_{nm}}.
    \end{align}
    Otherwise, the contraction constant is bounded by
    \begin{align}
        \label{equality:rho-of-ios-2}
        \rho \le& \max_{n\in\RR}  \frac{15\sum_{m \in \cH_n}w'_{nm}}{1-3\sum_{m \in \cH_n}w'_{nm}}.
    \end{align}
\end{lemma}

Note that when the subgraph of all the non-Byzantine agents is connected, the virtual mixing matrix $W$ associated with trimmed mean satisfies $\lambda := 1-\|(I-\frac{1}{\vert \RR \vert}\bm{1}\bm{1}^\top)W\|^2 \in (0,1]$. For SCC and IOS, such results also hold as long as $w'_{nm} \neq 0$ when $n$ and $m$ are neighbors or identical.

In the context of decentralized learning, contraction constants and virtual mixing matrices of trimmed mean and SCC have not been investigated.
The work of \cite{wu2022byzantine} gives the contraction constant and virtual mixing matrix of IOS, but only considers the scenario that the Byzantine messages are not entirely removed. Our upper bound of the contraction constant is tighter since we also consider the scenario that the Byzantine messages are all successfully removed.
Our analysis incorporates new technical tools and applies to arbitrary topologies.


A contraction inequality similar to \eqref{inequality:robustness-of-aggregation-local} can be found in \cite{farhadkhani2022byzantine}, but is limited to federated learning on the star topology. The analysis does not involve the virtual mixing matrix $W$, which is crucial in decentralized learning.

{\color{black}
\begin{remark}
The above analysis reveals the ratio of Byzantine agents tolerable by each robust aggregation rule. Such a ratio is topology-dependent. (i) In trimmed mean, for each non-Byzantine agent $n$, the number of its non-Byzantine neighbors is at least twice of that of its Byzantine neighbors, namely, $|\mathcal{N}_n|\geq2|\mathcal{B}_n|$. (ii) In SCC, for each non-Byzantine agent $n$, the best threshold $\tau_n$ is related to the sums of the weights assigned to its non-Byzantine neighbors $\sum_{m \in \mathcal{R}_n} w'_{nm}$ and to its Byzantine neighbors $\sum_{m \in \mathcal{B}_n} w'_{nm}$. (iii) In IOS, for each non-Byzantine agent $n$, the sum of the largest $q_n$ weights assigned to its neighbors must be less than $\frac{1}{3}$, that is to say, $\sum_{m \in \cH_n}w'_{nm}< \frac{1}{3}$.
\end{remark}
}

\begin{table}[!htbp]
    \caption{Robust aggregation rules and their corresponding $\rho$ and $W$ on a fully connected topology}
    \label{table:rho-complete}
    \centering
    \begin{tabular}{cccc}
    \hline
    rule  &$\rho$  &$W$ \\
    \hline
    trimmed mean \cite{fang2022bridge} &  $ \frac{2|\B|}{\vert \RR \vert }\cdot \min\{D^{\frac{1}{2}},|\RR| ^{\frac{1}{2}}\}$  & $\frac{1}{\vert \RR \vert}\bm{1}\bm{1}^\top$ \\
    SCC \cite{he2022byzantine}  &  $\frac{4\sqrt{|\RR||\B|}}{|\N|}$ &$\frac{1}{\vert \N \vert}\bm{1}\bm{1}^\top+\frac{|\B|}{\vert \N \vert}\bm{I}$ \\
    IOS \cite{wu2022byzantine}  & $ \frac{|\B|}{\vert \RR \vert }$ &$\frac{1}{\vert \N \vert}\bm{1}\bm{1}^\top+\frac{|\B|}{\vert \N \vert}\bm{I}$\\
    \hline
    \end{tabular}
\end{table}

\subsection{Contraction Constants and Virtual Mixing Matrices on Fully Connected Topology: A Case Study}


As a case study, we investigate the contraction constants and virtual mixing matrices of different robust aggregation rules on a fully connected topology comprising of $|\RR|$ non-Byzantine agents and $|\B|$ Byzantine agents, with $|\RR| > |\B|$.
The corresponding contraction constants and virtual mixing matrices are presented in Table \ref{table:rho-complete}, based on Lemmas \ref{tm-rho}--\ref{ios-rho}. Therein, the assigned weights $w^\prime_{mn} = \frac{1}{|\mathcal{N}|}$ for SCC and IOS.


Observe that the contraction constant of trimmed mean is larger than ${2|\B|}/{|\RR|}$. Since $|\RR| > |\B|$, that of SCC is also larger than ${2|\B|}/{|\RR|}$. In contrast, the contraction constant of IOS is ${|\B|}/{|\RR|}$ and thus the smallest among the three robust aggregation rules. In addition, the contraction constant of trimmed mean has poor dependence on the model dimension or the number of non-Byzantine agents, as illustrated by the coefficient $\min\{D^{\frac{1}{2}},|\RR| ^{\frac{1}{2}}\}$.

In the next section, we will explore how the contraction constants and virtual mixing matrices of the three robust aggregation rules affect the tradeoff between privacy preservation and Byzantine-robustness, on general topologies.

\section{Numerical Experiments}
\label{s7}

By default, we construct a random Erdos-R{\'e}nyi graph of $12$ agents, and let each pair of agents be connected with probability 0.7. Among them, we randomly select $|\mathcal{B}| = 2$ to be Byzantine and the rest $|\RR|$ $= 10$ to be non-Byzantine.
 {We also conduct additional numerical experiments on a network of $|\mathcal{B}| = 30$ Byzantine $|\RR|$ $= 70$ non-Byzantine agents to show the scalability of the proposed method; see Section \ref{app-d}.}
We use the MNIST dataset, which contains 10 handwritten digits from 0 to 9, with 60,000 training images and 10,000 testing images.
{\color{black}We consider both the i.i.d. and non-i.i.d. settings.} In the i.i.d. setting, the training images are randomly and evenly allocated to all non-Byzantine agents. In the non-i.i.d. setting, each non-Byzantine agent only possesses the training images of one handwritten digit, such that the local data distributions are different.
The overall aggregated cost is softmax regression with a strongly convex regularization term.
The code is available online\footnote{\url{https://github.com/haoxiangye/ppbrl}}.

In the numerical experiments, we implement the proposed privacy-preserving and Byzantine-robust decentralized SGD with different robust aggregation rules: trimmed mean \cite{fang2022bridge}, SCC \cite{he2022byzantine} and IOS \cite{wu2022byzantine}. The theoretical step size $\alpha^{k} = O(\frac{1}{k})$ diminishes too fast. Therefore, we set $\alpha^k = \frac{0.9}{\sqrt{k}}$ instead as in \cite{wu2022byzantine}.
{\color{black} We set the clipping threshold $\tau = 1$ as in \cite{Allouah:302948,li2024clipped,NEURIPS2020_9ecff545}.}
The parameters $q_n$ in trimmed mean and IOS and the parameters $\tau_n$ in SCC are tuned to the best.
Performance metrics are accuracy of the non-Byzantine agents' average model $\bar{\vx}^k$ and disagreement measure $H^k$.
We test the following classical Byzantine attacks.

\noindent\textbf{Gaussian attacks \cite{ma2022differentially}.} The Byzantine agents send messages with elements following {\color{black}Gaussian distribution $N(0,1)$.} \\
\noindent\textbf{Sign-flipping attacks \cite{december}.} To non-Byzantine agent $n$, its Byzantine neighbors multiply $\frac{1}{|\RR_n|}\sum_{m\in \RR_n }\tilde\vx_{m,m}^{k}$ with {\color{black}a negative constant $-1$} and send the result, in the $k$-th iteration.\\
\noindent\textbf{Isolating attacks \cite{wu2022byzantine}.} To non-Byzantine agent $n$, its Byzantine neighbors send ${(\tilde\vx_{n,n}^{k}-\sum_{m \in \RR_n}w'_{nm} \tilde\vx_{m,m}^{k})}/ \sum_{m \in \B_n}$ $w'_{nm}$ such that the weighted sum of its received messages becomes $\tilde\vx_{n,n}^{k}$ in the $k$-th iteration of decentralized SGD. This is equivalent to that non-Byzantine agent $n$ is isolated.\\
 {\noindent\textbf{AGR attacks \cite{shejwalkar2021manipulating}.} To non-Byzantine agent $n$, its Byzantine neighbors subtract $\frac{1}{|\RR_n|}\sum_{m\in \RR_n }\tilde\vx_{m,m}^{k}$ by the sign and send the result, in the $k$-th iteration.\\
\noindent\textbf{ALIE attacks\cite{baruch2019little}.} To non-Byzantine agent $n$, its Byzantine neighbors send $\frac{1}{|\RR_n|}\sum_{m\in \RR_n }\tilde\vx_{m,m}^{k} + r_n^k \Delta_n^k$ in the $k$-th iteration, where $\Delta_n^k$ represents the coordinate-wise standard deviation of $\{\tilde\vx_{m,m}^{k}\}_{m\in \RR_n}$ and $r_n^k$ is the scale factor.}

\subsection{Tradeoffs in Different Robust Aggregation Rules}

\begin{figure*}
\centering
{\includegraphics[width=7in]{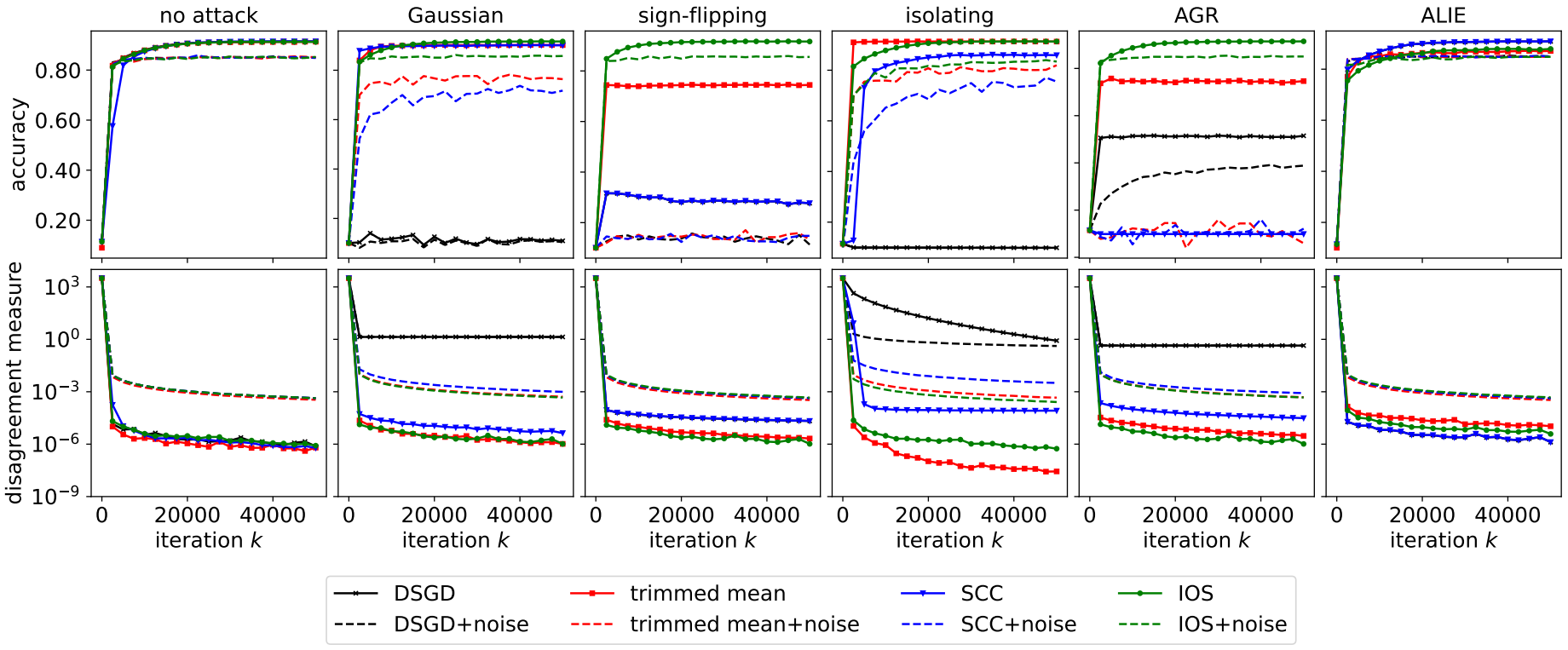}}
\caption{Different robust aggregation rules in non-i.i.d. setting on MNIST.}\label{noniid}
\end{figure*}

\begin{figure*}[h]
\centering
{\includegraphics[width=7in]{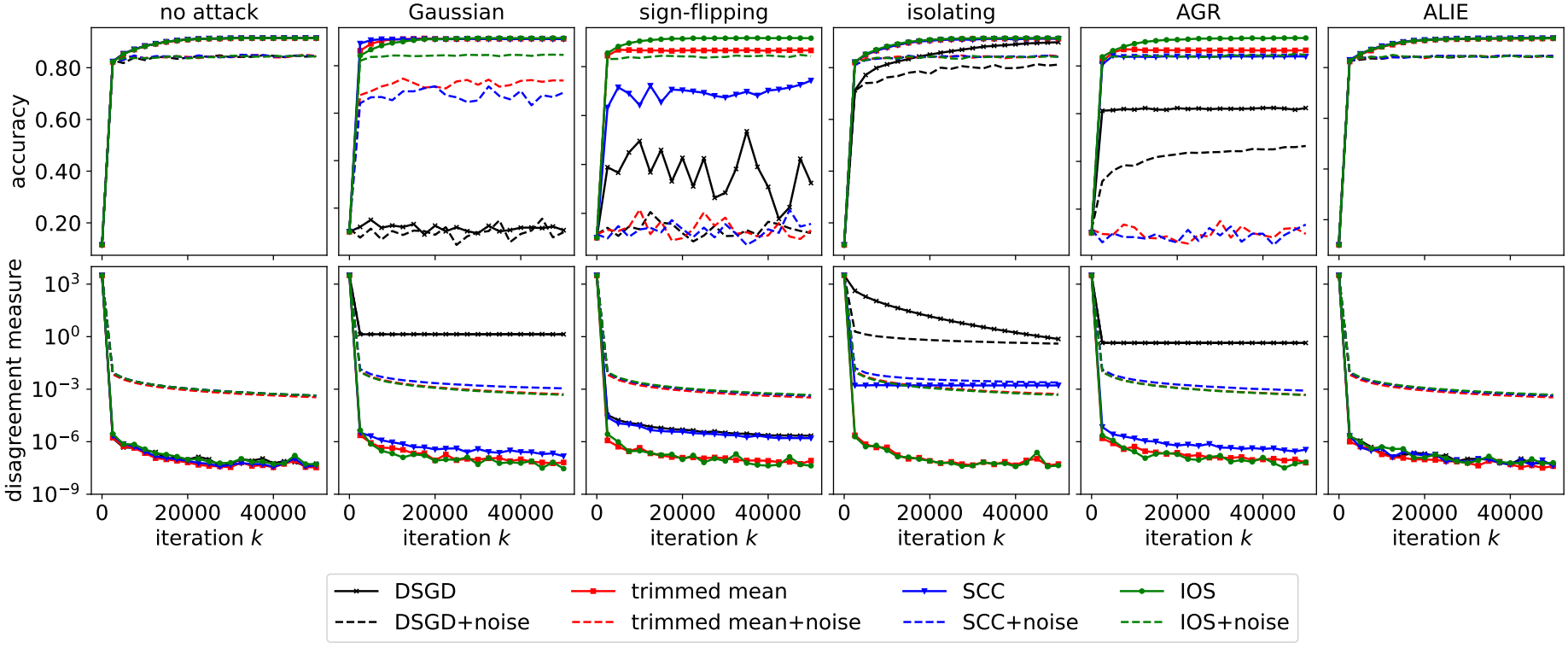}}
\caption{Different robust aggregation rules in i.i.d. setting on MNIST.}\label{iid}
\end{figure*}

We demonstrate the performance of different robust aggregation rules, {\color{black} with a fixed noise level $C=0.2$ such that the total privacy budgets are $(\epsilon,\delta)=(4.72,10^{-4})$.}
 {The values of $S$, $B$ and $K$ are $6000$, $32$ and $50000$, respectively.}
Privacy-preserving decentralized SGD (DSGD), which is not Byzantine-robust, is used as a baseline. We also run the noise-free methods as comparisons.
The results in the non-i.i.d. setting are depicted in Fig. \ref{noniid}, and the results in the i.i.d. setting can be found in 
Fig. \ref{iid}.

When no Byzantine agents are present (by deleting the two Byzantine agents and their associated links), all the methods work well. This is because the associated contraction constants $\rho$ are close to zero, while the virtual mixing matrices $W$ have $\chi^2$ close to zero. The noise added for privacy preservation does not affect the performance too much as our analysis predicts.

{\color{black}Under Gaussian, isolating and ALIE attacks,} all the robust aggregation rules largely outperform the vulnerable privacy-preserving DSGD, but have degraded performance compared to the attack-free case. The noise added for privacy preservation exacerbates the performance degradation, which also matches our analysis. This phenomenon confirms the tradeoff between privacy preservation and Byzantine-robustness.

{\color{black}Meanwhile, sign-flipping and AGR attacks turn out to be the strongest in the numerical experiments} -- under them trimmed mean and SCC become unstable without noise, and fail with noise.
IOS performs well in both cases. We conjecture that it is because IOS is designed to have small $\rho$ and $\chi^2$ \cite{wu2022byzantine}.
To validate this conjecture, we present Table \ref{table:rho-case1} that demonstrates $\rho^2 + \chi^2$ for different robust aggregation rules. IOS has the smallest $\rho^2+\chi^2$, resulting in the least performance degradation when combined with the Gaussian noise mechanism. In contrast, SCC has the largest $\rho^2+\chi^2$, and thus suffers from severe performance degradation. This observation verifies our theoretical finding that a robust aggregation rule with small values of $\rho$ and $\chi^2$ yields a favorable tradeoff between privacy preservation and Byzantine-robustness. With them and given a target learning error, the added noise level can be high, which is beneficial to privacy preservation.



\begin{table}[!htbp]
    \caption{Robust aggregation rules and the corresponding upper bounds of $\rho^2 + \chi^2$}
    \label{table:rho-case1}
    \centering
    \begin{tabular}{cccc}
    \hline
    rule  &$\rho^2$  &$\chi^2$ & $\rho^2+\chi^2$ \\
    \hline
    trimmed mean \cite{fang2022bridge} & 1.9592 & 0.1018  &2.0610 \\
    SCC \cite{he2022byzantine}  & 2.7654 & 0 & 2.7654\\
    IOS \cite{wu2022byzantine}  & 0.1111 & 0 & 0.1111\\
    \hline
    \end{tabular}
\end{table}

\begin{figure*}
\centering
{\includegraphics[width=7in]{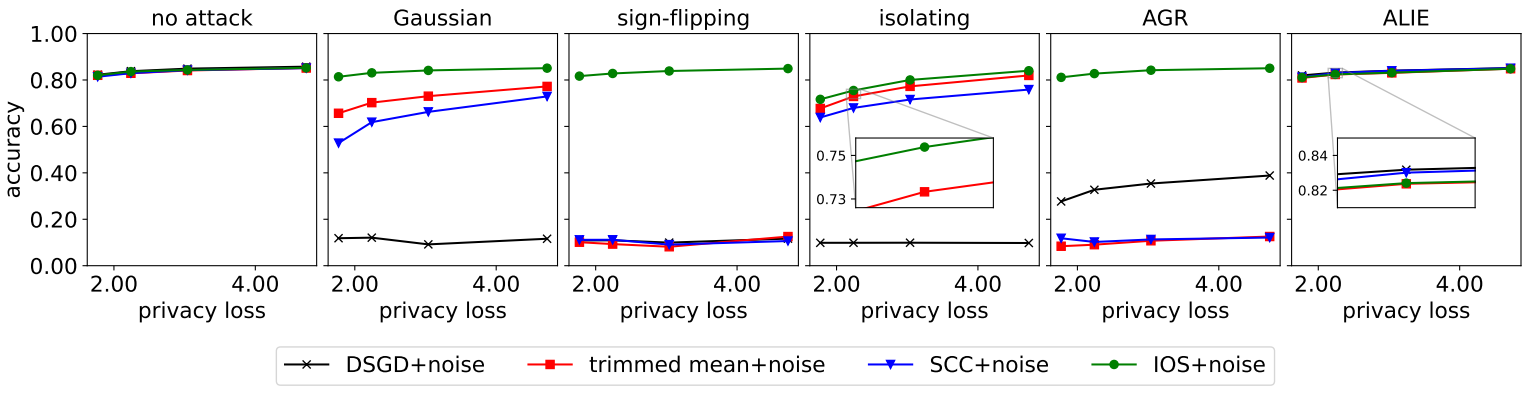}}
\caption{Different robust aggregation rules under different noise levels in non-i.i.d. setting on MNIST.}\label{compare-noniid}
\end{figure*}

\subsection{Tradeoffs in Different Noise Levels}
\begin{figure*}[h]
\centering
{\includegraphics[width=7in]{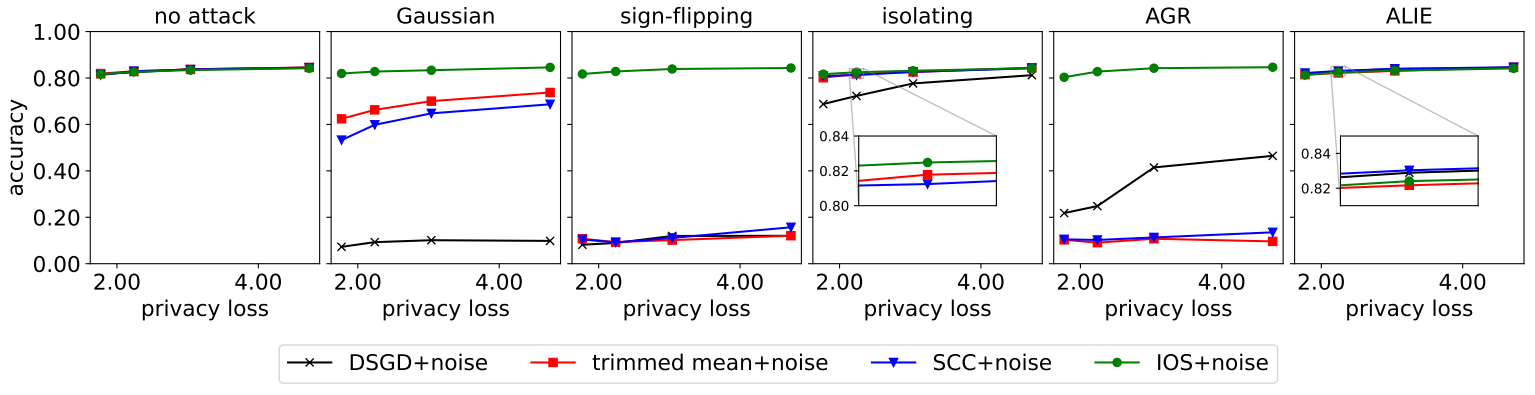}}
\caption{Different robust aggregation rules under different noise levels in i.i.d. setting on MNIST.}\label{compare-iid}
\end{figure*}

To further illustrate the tradeoff between privacy preservation and Byzantine-robustness, we test different noise levels for different robust aggregation rules in the non-i.i.d setting, as shown in Fig. \ref{compare-noniid}.
The results in the i.i.d setting are shown in Fig. \ref{compare-iid}.
{\color{black}We set the noise levels to $C = 0.2$, $0.3$, $0.4$, $0.5$ and the total privacy budgets after $K=50000$ iterations are $(\epsilon,\delta)$ $=(4.72,10^{-4})$, $(\epsilon,\delta)=(3.04,10^{-4})$, $(\epsilon,\delta)=(2.24,10^{-4})$, $(\epsilon,\delta)=(1.77,10^{-4})$, respectively.}
We can observe a clear tradeoff: although increasing the level of noise helps privacy preservation, it hurts Byzantine-robustness, yielding a lower accuracy.
Again, the results in Fig. \ref{compare-noniid} show that a robust aggregation rule with small values of $\rho$ and $\chi^2$ yields a favorable tradeoff.



\begin{figure*}
\centering
{\includegraphics[width=7in]{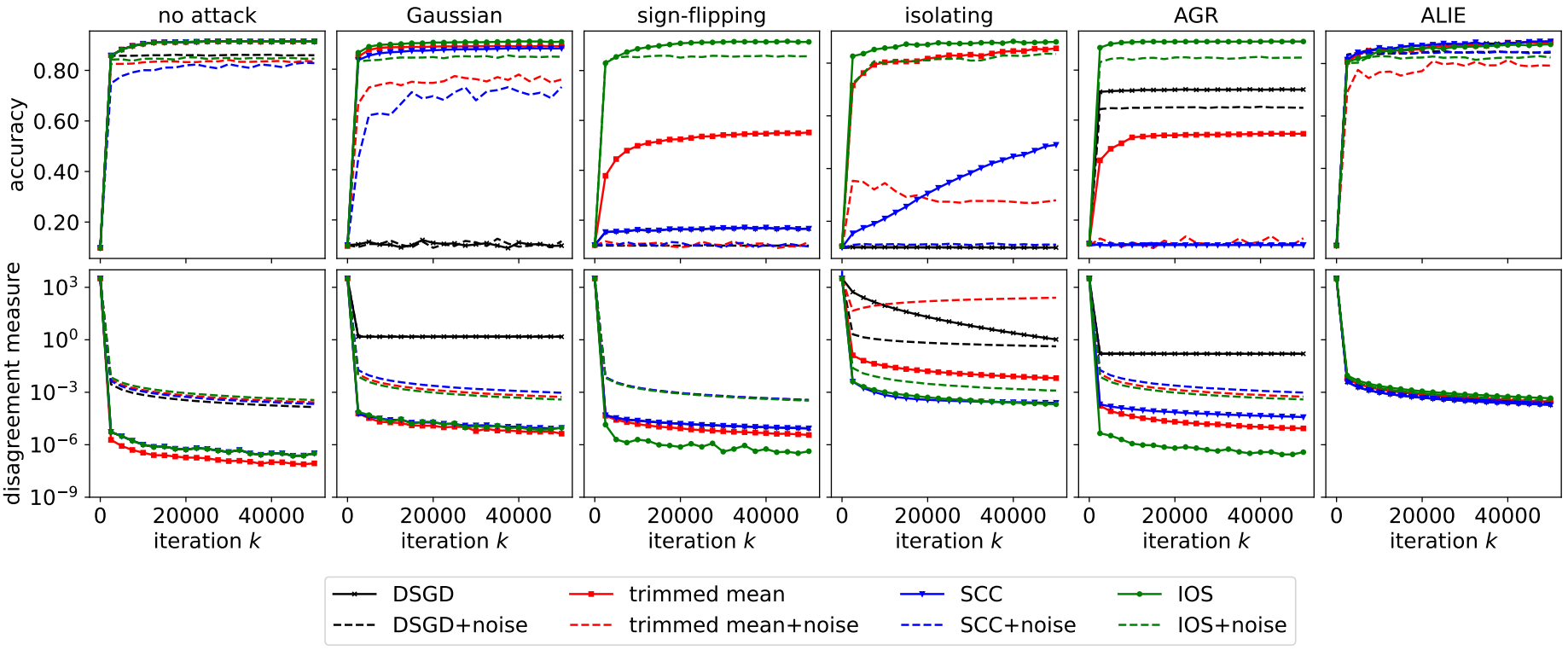}}
\caption{Different robust aggregation rules in non-i.i.d. setting on MNIST for the network of 100 agents.}\label{noniid-100}
\end{figure*}

\subsection{Performance under DLG Attacks}
\begin{figure}[h]
\centering
{\includegraphics[width=3.5in]{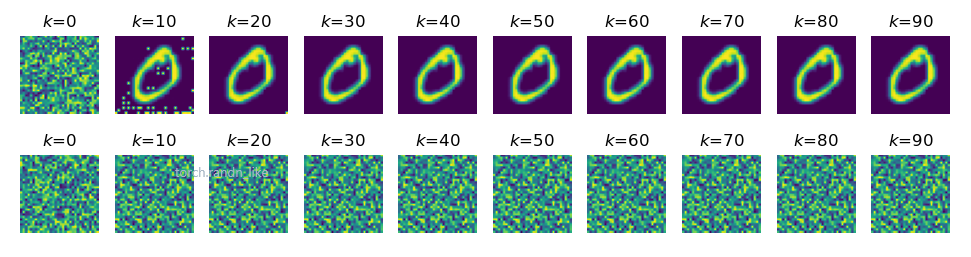}}
\caption{DLG recovery for Byzantine-robust decentralized SGD (TOP) versus privacy-preserving and Byzantine-robust decentralized SGD (BOTTOM).}\label{dlg}
\end{figure}

To depict the privacy preservation ability of the proposed method, we conduct the \textbf{DLG (deep leakage from gradients) attacks}. Following the setting of \cite{Zhu2019}, we consider the worst case. Let an agent be surrounded by curious agents so that its received and transmitted messages are all exposed -- in this case, for any prescribed robust aggregation rule the curious agents can always know the output. Also let the chosen agent compute its stochastic gradients from the same sample of digit ``0". The goal of the curious agents is to recover this sample. Fig. \ref{dlg} illustrates that the curious agents effectively recover the sample for Byzantine-robust decentralized SGD, but fail when the Gaussian noise mechanism is incorporated.

{\color{black}
\subsection{Performance on a Larger Network}
\label{app-d}
We construct a random Erdos-R{\'e}nyi graph of $100$ agents, and let each pair of agents be connected with probability 0.7. Among them, we randomly select $|\mathcal{B}| = 30$ to be Byzantine and the rest $|\RR|$ $= 70$ to be non-Byzantine.
We use the MNIST dataset and consider the non-i.i.d. data distribution.
Fig. \ref{noniid-100} demonstrates the performance of different robust aggregation rules, with a fixed noise level $C=0.2$ such that the total privacy budgets are $(\epsilon,\delta)=(53.44,10^{-4})$. The observations are consistent with those on the smaller network. Under various attacks, all robust aggregation rules show degraded performance compared to the attack-free case. The noise added for privacy preservation exacerbates this performance degradation, confirming the tradeoff between privacy preservation and Byzantine-robustness. The better performance of IOS indicates that a robust aggregation rule with small values of $\rho$ and $\chi^2$ yields a favorable tradeoff; see Table \ref{table:rho-100}.
}

\begin{table}[!htbp]
    \caption{Robust aggregation rules and the corresponding upper bounds of $\rho^2 + \chi^2$ for the network of 100 agents}
    \label{table:rho-100}
    \centering
    \begin{tabular}{cccc}
    \hline
    rule  &$\rho^2$  &$\chi^2$ & $\rho^2+\chi^2$ \\
    \hline
    trimmed mean \cite{fang2022bridge} & 2.8348 & 0.0703  & 2.9051\\
    SCC \cite{he2022byzantine}  & 3.6344 & 0 & 3.6344\\
    IOS \cite{wu2022byzantine}  & 0.2870 & 0 & 0.2870\\
    \hline
    \end{tabular}
\end{table}

\section{Conclusions}
\label{s8}
In this paper, we jointly consider privacy preservation and Byzantine-robustness in decentralized learning.
We study a generic privacy-preserving and Byzantine-robust decentralized SGD framework,
where the Gaussian noise mechanism is employed to protect privacy and a robust aggregation rule is used to counteract Byzantine attacks.
We discover an essential tradeoff between additive noise-based privacy preservation and Byzantine-robustness, and provide guidelines for achieving a favorable tradeoff -- the chosen robust aggregation rule should be associated with a small contraction factor and a doubly stochastic virtual mixing matrix.
These preliminary results are expected to motivate more explorations in this interesting and important area.




\bibliographystyle{IEEEtran}
\bibliography{TDSC}

\newpage
\twocolumn[
\begin{@twocolumnfalse}
	\section*{\centering{Supplementary Material for \\ \emph{On the Tradeoff between Privacy Preservation and Byzantine-Robustness in Decentralized Learning	\\[15pt]}}}
\end{@twocolumnfalse}
]
\setcounter{section}{0}

\section{Proof of Theorem \ref{t5}}
\label{proof-a}
\begin{proof}
   For convenience, define $X^{k} = [\vx^{k}_1, \cdots, \vx^{k}_{|\RR|}]^\top \in$ $\mathbb{R}^{|\RR| \times D}$ and $\tilde{X}^{k} = [\tilde\vx^{k}_{1,1}, \cdots, \tilde\vx^{k}_{|\RR|, |\RR|}]^\top \in \mathbb{R}^{|\RR| \times D}$. For any $u \in (0, 1)$, it holds
    \begin{align}
        \label{inequality:ce-itearation-0}
        &  \sum_{n\in \RR}   \|\vx^{k+1}_{n}-\bar\vx^{k+1}\|^2   \\
        = & \|(I-\frac{1}{\vert \RR \vert}\bm{1}\bm{1}^\top) X^{k+1}\|_{F}^2  \nonumber\\
         \le & \frac{1}{1-u} \|(I-\frac{1}{\vert \RR \vert}\bm{1}\bm{1}^\top)W \tilde{X}^{k}\|_{F}^2
        + \frac{2}{u} \|X^{k+1}-W\tilde{X}^{k}\|_{F}^2
        \nonumber \\
        &
        + \frac{2}{u} \|\frac{1}{\vert \RR \vert}\bm{1}\bm{1}^\top X^{k+1}-\frac{1}{\vert \RR \vert}\bm{1}\bm{1}^\top W\tilde{X}^{k}\|_{F}^2,  \nonumber
    \end{align}
    where the equality is due to $\bar\vx^{k+1} := \frac{1}{|\RR|}\sum_{n\in \RR}\vx^{k+1}_n$ and the inequality comes from $\|\bm{a}+\bm{b}+\bm{c}\|^2\le \frac{1}{1-u}\|\bm{a}\|^2+\frac{2}{u}\|\bm{b}\|^2+\frac{2}{u}\|\bm{c}\|^2$.

    For the first-term at the right-hand side (RHS) of \eqref{inequality:ce-itearation-0}, we have
    \begin{align}
        \label{inequality:ce-itearation-0-1}
        & \|(I-\frac{1}{\vert \RR \vert}\bm{1}\bm{1}^\top) W\tilde{X}^{k}\|_{F}^2
         \\
        =&  \|(I-\frac{1}{\vert \RR \vert}\bm{1}\bm{1}^\top)W (I-\frac{1}{\vert \RR \vert}\bm{1}\bm{1}^\top) \tilde{X}^{k}\|_{F}^2
        \nonumber \\
        \le&  \|(I-\frac{1}{\vert \RR \vert}\bm{1}\bm{1}^\top) W\|^2\|(I-\frac{1}{\vert \RR \vert}\bm{1}\bm{1}^\top) \tilde{X}^{k}\|_{F}^2
        \nonumber \\
        =& (1-\lambda)  \| (I-\frac{1}{\vert \RR \vert}\bm{1}\bm{1}^\top) \tilde{X}^{k}\|_{F}^2.  \nonumber
    \end{align}
    For the second term at the RHS of \eqref{inequality:ce-itearation-0}, we use the contractive property of robust aggregation rules $ \{ \A_n \}_{n \in  \RR}$ in \eqref{inequality:robustness-of-aggregation-local} to derive
    \begin{align}
        \label{inequality:ce-itearation-0-2}
        & \|X^{k+1}-W\tilde{X}^{k}\|_{F}^2
          \\
        =&  \sum_{n\in \RR}  \|\A_n (\tilde\vx^{k}_{n,n}, \{\tilde{\vx}^{k}_{m,n}\}_{m\in \RR_n\cup\B_n})-\hat\vx^{k}_n\|^2
        \nonumber \\
        \le& \rho^2\sum_{n\in \RR} \max_{m\in \RR_n \cup \{n\}}\| \tilde{\vx}^{k}_{m,n}-\hat\vx^{k}_n\|^2
        \nonumber \\
        \le& 4\rho^2 \sum_{n\in \RR}  \max_{m\in \RR}\| \tilde{\vx}^{k}_{m,m}-\hat\vx^{k}\|^2  \nonumber \\
        \le& 4\rho^2 \vert \RR \vert \sum_{n\in \RR}  \| \tilde{\vx}^{k}_{n,n}-\hat\vx^{k}\|^2 \nonumber \\
        = & 4\rho^2 \vert \RR \vert  \|(I-\frac{1}{\vert \RR \vert}\bm{1}\bm{1}^\top) \tilde{X}^{k}\|_{F}^2,  \nonumber
    \end{align}
where $\hat\vx^{k}_n := \sum_{m \in \RR_n\cup\{n\}}w_{nm} \tilde\vx^{k}_{m,n}$ is the weighted average and $\hat \vx^{k} :=  \frac{1}{|\RR|}\sum_{n\in \RR}\tilde\vx_{n,n}^{k}$ is the average. The second inequality holds true because
\begin{align}
\label{ineq-ce4}
		& \max_{m\in  \RR_n \cup \{n\}}  \|\tilde{\vx}^{k}_{m,m}-\hat\vx^{k}_n\|^2
		  \\
		\le& 2  \max_{m\in  \RR_n \cup \{n\}}  \|\tilde{\vx}^{k}_{m,m}-\hat\vx^{k}\|^2
		+2  \|\hat\vx^{k}- \hat\vx^{k}_n\|^2
		\nonumber \\
		\le& 2   \max_{m\in  \RR_n \cup \{n\}}  \|\tilde{\vx}^{k}_{m,m}-\hat\vx^{k}\|^2
		+2 \max_{n\in\RR}\|\tilde{\vx}^{k}_{n,n}-\hat\vx^{k}\|^2 \nonumber \\
            \le& 4 \max_{n\in\RR}\|\tilde{\vx}^{k}_{n,n}-\hat\vx^{k}\|^2.  \nonumber
	\end{align}
    A similar technique can be applied to the third term at the RHS of \eqref{inequality:ce-itearation-0}, yielding
    \begin{align}
        \label{inequality:ce-itearation-0-3}
        &\E\|\frac{1}{\vert \RR \vert}\bm{1}\bm{1}^\top X^{k+1}-\frac{1}{\vert \RR \vert}\bm{1}\bm{1}^\top W\tilde{X}^{k}\|_{F}^2 \\
        \leq &   \|\frac{1}{\vert \RR \vert}\bm{1}\bm{1}^\top\|^2 \| X^{k+1} - W\tilde{X}^{k}\|_{F}^2   \nonumber \\
        \leq & 4\rho^2 |\RR|   \|(I-\frac{1}{\vert \RR \vert}\bm{1}\bm{1}^\top) \tilde{X}^{k}\|_{F}^2.   \nonumber
    \end{align}
Substituting \eqref{inequality:ce-itearation-0-1}--\eqref{inequality:ce-itearation-0-3} back into \eqref{inequality:ce-itearation-0} and taking expectation over all random variables up to iteration $k+1$, we have
    \begin{align}
    \label{eq:A6}
            &\sum_{n\in \RR} \E \|\vx^{k+1}_{n}-\bar\vx^{k+1}\|^2   \\
        \le & \lp \frac{1-\lambda}{1-u}
        + \frac{16\rho^2\vert \RR \vert}{u} \rp \E \|(I-\frac{1}{\vert \RR \vert}\bm{1}\bm{1}^\top)\tilde{X}^{k}\|_{F}^2.   \nonumber
    \end{align}

    Now we consider
    \begin{align}
     \label{ineq2-0}
     \E\|(I-\frac{1}{\vert \RR \vert}\bm{1}\bm{1}^\top) \tilde{X}^{k}\|_{F}^2  = \sum_{n\in\RR} \E \|\tilde{\vx}^{k}_{n,n}-\hat\vx^{k}\|^2.
    \end{align}
{\color{black} For notational convenience, here we define $\varphi = \min\{M,\tau\}$. Because $\tilde\vx^{k}_{n,n} =\vx_n^k - \alpha^k \tilde\nabla f_n(\vx^{k}_n; \xi_n^{k}) + \mathbf{e}_n^{k}$ and $\hat\vx^{k} = \bar\vx^k -  \frac{\alpha^k}{\vert \RR \vert} \sum_{m\in\RR} \tilde\nabla f_m(\vx^{k}_m; \xi_m^{k}) + \frac{1}{\vert \RR \vert} \sum_{m\in\RR} \mathbf{e}_m^{k}$, for any $v \in (0,1)$ we have
    \begin{align}
    \label{ineq2}
    &\sum_{n\in\RR} \E \|\tilde{\vx}^{k}_{n,n}-\hat\vx^{k}\|^2   \\
        \leq & \frac{1}{1-v} \sum_{n\in\RR}  \E \|\vx^k_{n} - \bar\vx^k \|^2 + \frac{1}{v} \sum_{n\in\RR} \E \|\ve^{k}_{n} - \frac{1}{\vert \RR \vert} \sum_{m\in\RR} \mathbf{e}_m^{k} \|^2  \nonumber\\
        & + \frac{(\alpha^k)^2}{v} \sum_{n\in\RR} \E \| \tilde\nabla f_n(\vx^{k}_n; \xi_n^{k}) - \frac{1}{\vert \RR \vert} \sum_{m\in\RR} \tilde\nabla f_m(\vx^{k}_m; \xi_m^{k})  \|^2   \nonumber\\
        \le& \frac{1}{1-v}\sum_{n\in\RR} \| \vx^{k}_{n}-\bar\vx^{k}\|^2
        +\frac{4\vert \RR \vert(\alpha^k)^2\varphi^2+D\vert \RR \vert(\sigma^k)^2}{v}.  \nonumber
    \end{align}}
Here the last inequality holds true because
\begin{align}
    &\E \|\ve^{k}_{n} - \frac{1}{\vert \RR \vert} \sum_{m\in\RR} \mathbf{e}_m^{k} \|^2 \\
    =& \E \|(1- \frac{1}{\vert \RR \vert})\ve^{k}_{n} - \frac{1}{\vert \RR \vert} \sum_{m\in\RR \setminus \{n\}} \mathbf{e}_m^{k} \|^2  \nonumber \\
    =& (1- \frac{1}{\vert \RR \vert})^2 D (\sigma^k)^2 + \frac{|\RR|-1}{\vert \RR \vert^2} D(\sigma^k)^2  \nonumber\\
    \leq & D(\sigma^k)^2. \nonumber
\end{align}

    Applying (\ref{ineq2}) to \eqref{eq:A6}, we have
    \begin{align}
        \label{inequality:ce-itearation-1}
         & \E H^{k+1}
        = \frac{1}{\vert \RR \vert} \E\|(I-\frac{1}{\vert \RR \vert}\bm{1}\bm{1}^\top) X^{k+1}\|_{F}^2  \\
        \leq & \frac{1}{\vert \RR \vert}  \lp \frac{1-\lambda}{1-u}
        + \frac{16\rho^2\vert \RR \vert}{u} \rp \E \|(I-\frac{1}{\vert \RR \vert}\bm{1}\bm{1}^\top)\tilde{X}^{k}\|_{F}^2 \nonumber \\
        \leq &   \lp \frac{1-\lambda}{1-u}
        + \frac{16\rho^2\vert \RR \vert}{u} \rp \lp \frac{1}{1-v} \E H^k
        +\frac{4(\alpha^k)^2\varphi^2+D(\sigma^k)^2}{v}\rp  \nonumber \\
        \leq &   \lp 1-\lambda
        + 8\rho \sqrt{\vert \RR \vert} \rp \lp \frac{1}{1-v} \E H^k
        +\frac{4(\alpha^k)^2\varphi^2+D(\sigma^k)^2}{v}\rp  \nonumber \\
        \leq &   \lp 1- \omega_1 \rp \lp \frac{1}{1-v} \E H^k
        +\frac{4(\alpha^k)^2\varphi^2+D(\sigma^k)^2}{v}\rp,   \nonumber
    \end{align}
    where $\omega_1 := \lambda - 8\rho \sqrt{\vert \RR \vert}$. The last inequality comes from choosing $u=4\rho \sqrt{\vert \RR \vert} \leq \lambda$ and using the fact that $\frac{1-\lambda}{1-u}\le 1-\lambda+u$ for any $u \leq \lambda$.

    If we set $\frac{1}{1-v}=1+\omega_1$, \eqref{inequality:ce-itearation-1} becomes
    \begin{align}
        \label{inequality:ce-itearation-6-Hk-iteration}
        \hspace{-2em} \E H^{k+1} \leq \lp 1-\omega_1^2\rp \E H^k +   4\omega_2(\alpha^k)^2 \varphi^2 + D\omega_2(\sigma^k)^2,
    \end{align}
     where $\omega_2 := \frac{1-\omega_1^2}{\omega_1}$.
    We further choose the decaying step size  {\color{black}$\alpha^k= \frac{8}{\mu (k+k_0)}$} and set $\sigma^k =C \alpha^k$. Then, using telescopic cancellation on \eqref{inequality:ce-itearation-6-Hk-iteration} from $0$ to $k$, we deduce that
    \begin{align}
        &\E H^{k}
        \le  (1-\omega_1^2)^k H^0  \\
         &+\frac{256\omega_2 \varphi^2}{\mu^2}\lp \frac{1}{(k+k_0-1)^2}+\cdot\cdot\cdot+\frac{(1-\omega_1^2)^{k-1}}{(k_0)^2} \rp  \nonumber\\
        &+\frac{64\omega_2 C^2}{\mu^2} \lp \frac{1}{(k+k_0-1)^2}+\cdot\cdot\cdot+\frac{(1-\omega_1^2)^{k-1}}{(k_0)^2} \rp.  \nonumber
    \end{align}
    According to Lemma 5 in \cite{9462519}, there exists a certain constant $a_1 \geq \frac{(k_0+1)^2}{k_0^2}$ such that
     \begin{align}
            &\E H^{k}
        \le  (1-\omega_1^2)^k H^0
        +\frac{512 \omega_2 a_1 \varphi^2+128\omega_2 a_1 C^2}{\mu^2\omega_1^2} \frac{1}{(k+k_0)^2},  \nonumber
    \end{align}
    which completes the proof.
\end{proof}

\section{Proof of Theorem \ref{t4}}
\label{proof-b}
\begin{proof}
     Recall that $\bar\vx^{k+1} := \frac{1}{|\RR|}\sum_{n\in \RR}\vx^{k+1}_n$ is the average of all non-Byzantine local models at iteration $k+1$ and $\hat \vx^{k} :=  \frac{1}{|\RR|}\sum_{n\in \RR}\tilde\vx_{n,n}^{k}$ is the the average of all non-Byzantine local models with noise at iteration $k$. We have
    \begin{align}
    \label{in1}
            &\E \| \bar\vx^{k+1}-\vx^* \|^2  \\
            = &\E \Vert \frac{1}{\vert \RR \vert}\sum_{n\in  \RR} \A_n (\tilde\vx^{k}_{n,n}, \{\tilde{\vx}^{k}_{m,n}\}_{m\in \N_n})-\vx^* \Vert^2  \nonumber\\
            \leq& \frac{1}{1-a}\E \| \frac{1}{\vert \RR \vert}\sum_{n\in  \RR} \A_n (\tilde\vx^{k}_{n,n}, \{\tilde{\vx}^{k}_{m,n}\}_{m\in \N_n}) - \hat \vx^{k} \|^2  \nonumber\\
            &+ \frac{1}{a}\E \| \hat\vx^{k} - \vx^* \|^2  \nonumber\\
            \leq& \frac{2}{1-a}\E \| \frac{1}{\vert \RR \vert}\sum_{n\in  \RR} \lp \A_n (\tilde\vx^{k}_{n,n}, \{\tilde{\vx}^{k}_{m,n}\}_{m\in \N_n}) - \hat \vx_n^{k} \rp \|^2  \nonumber\\
            &+\frac{2}{1-a}\E \|  \frac{1}{\vert \RR \vert}\sum_{n\in  \RR}\hat \vx_n^{k} - \hat \vx^{k} \|^2+ \frac{1}{a}\E \| \hat\vx^{k} - \vx^* \|^2,   \nonumber
    \end{align}
    where $a \in (0,1)$ is a constant and the expectation is taken over all random variables up to iteration $k+1$. 

For the second term at the RHS of (\ref{in1}), we have
\begin{align}
\label{in2}
        &\E \| \frac{1}{\vert \RR \vert}\sum_{n\in  \RR} \hat \vx_n^{k} - \hat\vx^{k} \|^2  \\
        =& \E \lnorm \frac{1}{\vert \RR \vert} \bm{1}^\top (W\tilde{X}^{k}-\frac{1}{\vert \RR \vert}\bm{1}\bm{1}^{\top}\tilde{X}^{k})
        \rnorm^2
          \nonumber\\
        =& \frac{1}{\vert \RR \vert^2} \E \lnorm (\bm{1}^\top W-\bm{1}^{\top})
        (\tilde{X}^{k}-\frac{1}{\vert \RR \vert}\bm{1}\bm{1}^{\top}\tilde{X}^{k})
        \rnorm^2
         \nonumber\\
        \le& \frac{1}{\vert \RR \vert^2} \lnorm W^\top\bm{1} -\bm{1}
        \rnorm^2
       \E \lnorm \tilde{X}^{k}-\frac{1}{\vert \RR \vert}\bm{1}\bm{1}^{\top}\tilde{X}^{k}
        \rnorm^2_F
          \nonumber\\
		=& \frac{\chi^2}{\vert \RR \vert}\sum_{n\in\RR} \E \lnorm  \tilde{\vx}^{k}_{n,n}-\hat\vx^{k}\rnorm^2.   \nonumber
\end{align}

Applying Definition \ref{definition:mixing-matrix} and (\ref{in2}) to (\ref{in1}), we can obtain
\begin{align}
    \label{in3}
            &\E \| \bar\vx^{k+1}-\vx^* \|^2  \\
            \leq& \frac{2}{1-a} \frac{1}{\vert \RR \vert}\sum_{n\in  \RR} \rho^2 \E \max_{m\in  \RR_n\cup\{n\}}  \|\tilde{\vx}_{m,m}^{k} -  \hat \vx_n^{k}  \|^2  \nonumber\\
            &+ \frac{2 \chi^2}{(1-a)\vert \RR \vert}\sum_{n\in  \RR}\E \| \tilde{\vx}^{k}_{n,n}-\hat\vx^{k}\|^2 + \frac{1}{a}\E \| \hat\vx^{k} - \vx^* \|^2.   \nonumber
    \end{align}
According to (\ref{ineq-ce4}), we have
\begin{align}
\label{in5}
        &\E \| \bar\vx^{k+1}-\vx^* \|^2  \\
        \leq & \frac{1}{a}\E \| \hat\vx^{k} - \vx^* \|^2 +\frac{8\rho^2}{1-a}  \E\max_{n\in  \RR} \|\tilde{\vx}_{n,n}^{k} -  \hat\vx^{k}  \|^2  \nonumber\\
        &+ \frac{2 \chi^2}{(1-a)\vert \RR \vert}\sum_{n\in  \RR}\E \| \tilde{\vx}^{k}_{n,n}-\hat\vx^{k}\|^2   \nonumber
\end{align}

Due to the lack of Lipschitz continuous stochastic gradient, it is imperative to invoke Assumption \ref{assumption:convex} to establish a bound for $\E \| \hat\vx^{k}-\vx^* \|^2$, as
{\color{black}\begin{align}
\label{in6}
            &\E \| \hat\vx^{k}-\vx^* \|^2  \\
            \leq & \E \| \bar\vx^k-\vx^* \|^2 + (\alpha^k)^2\varphi^2  + \E\|\frac{1}{|\RR|} \sum_{m\in\RR} \mathbf{e}_m^{k}\|^2 \nonumber\\
            &- 2\alpha^k\E \langle \frac{1}{\vert \RR \vert}\sum_{n\in\RR} \tilde \nabla f_n(\vx^{k}_n; \xi_n^{k}),\bar\vx^k-\vx^* \rangle . \nonumber
    \end{align}
For notational convenience, now we define $\nabla f_n(\vx^{k}_n; \xi_n^{k}) = \frac{1}{B} \sum_{\xi \in \xi_n^k } \nabla f_n(\vx^{k}_n; \xi)$ as the averaged stochastic gradient without clipping. There exists a bias between $\tilde \nabla f_n(\vx^{k}_n; \xi_n^{k})$ and $\nabla f_n(\vx^{k}_n; \xi_n^{k})$ that we must handle. For the last term at the RHS of (\ref{in6}), we can divide it into two parts
\begin{align}
\label{in6-0-1}
&\E \langle \tilde \nabla f_n(\vx^{k}_n; \xi_n^{k}),\bar\vx^k-\vx^* \rangle  \\
= &  \E \langle \tilde \nabla f_n(\vx^{k}_n; \xi_n^{k}) -  \nabla f_n(\vx^{k}_n; \xi_n^{k}), \bar\vx^k-\vx^* \rangle  \nonumber \\
 & + \E \langle \nabla f_n(\vx^{k}_n; \xi_n^{k})  ,\bar\vx^k-\vx^* \rangle.  \nonumber
\end{align}
For the first term at the RHS of (\ref{in6-0-1}),
\begin{align}
\label{in6-0-2}
\hspace{-1em} &  -\E \langle \tilde \nabla f_n(\vx^{k}_n; \xi_n^{k}) -  \nabla f_n(\vx^{k}_n; \xi_n^{k}), \bar\vx^k-\vx^* \rangle   \\
\hspace{-1em} \leq &  \E\|\tilde \nabla f_n(\vx^{k}_n; \xi_n^{k}) -  \nabla f_n(\vx^{k}_n; \xi_n^{k})\|  \|\bar\vx^k-\vx^*\| \nonumber \\
\hspace{-1em} = & \E \lnorm \frac{1}{B} \sum_{\xi \in \xi_n^k } \left[ Clip(\nabla f_n(\vx^{k}_n; \xi),\tau) - \nabla f_n(\vx^{k}_n; \xi) \right]\rnorm \|\bar\vx^k-\vx^*\| \nonumber \\
\hspace{-1em} =& \E \lnorm \frac{1}{B} \sum_{\xi \in \xi_n^k } \left(\min\{1,\frac{\tau}{\|\nabla f_n(\vx^{k}_n; \xi)\|}\}-1 \right) \nabla f_n(\vx^{k}_n; \xi) \rnorm \|\bar\vx^k-\vx^*\| \nonumber \\
\hspace{-1em} \leq &  \max\{M-\tau,0\}\E \|\bar\vx^k-\vx^*\|  \nonumber \\
\hspace{-1em} \leq & \frac{\mu}{8}\E\|\bar\vx^k-\vx^*\|^2 +\frac{8\max\{M-\tau,0\}^2}{\mu}.  \nonumber
\end{align}
where the last inequality holds true because $xy \leq ax^2+\frac{y^2}{a}$ and set $a=\frac{\mu}{8}$.
For the second term at the RHS of (\ref{in6-0-1}),
}
\begin{align}
\label{in6-1}
&\E \langle \nabla f_n(\vx^{k}_n; \xi_n^{k}),\bar\vx^k-\vx^* \rangle  \\
= &  \E \langle \nabla f_n(\vx^{k}_n; \xi_n^{k}),\bar\vx^k-\vx^k_n \rangle + \E \langle \nabla f_n(\vx^{k}_n; \xi_n^{k}),\vx^k_n-\vx^* \rangle  \nonumber \\
\geq & - M \E \| \bar\vx^k-\vx^k_n \| + \E \langle \nabla f_n(\vx^{k}_n; \xi_n^{k}),\vx^k_n-\vx^* \rangle .  \nonumber
\end{align}
With Assumption \ref{assumption:convex}, we can bound the second term at the RHS of (\ref{in6-1}) by
\begin{align}
\label{in6-2}
& \E \langle \nabla f_n(\vx^{k}_n; \xi_n^{k}),\vx^k_n-\vx^* \rangle \\
\geq & \E [ f_n(\vx_n^k) -  f_n(\vx^*) ] + \frac{\mu}{2} \E \|\vx_n^k-\vx^*\|^2  \nonumber \\
= &  \E [ f_n(\vx_n^k) - f_n(\bar\vx_k) ] + \E [ f_n(\bar\vx_k) - f_n(\vx^*) ]
+ \frac{\mu}{2} \E \|\vx_n^k-\vx^*\|^2 \nonumber \\
\geq & \E \langle \nabla f_n(\bar \vx^{k}; \xi_n^{k}),\vx^k_n-\bar \vx^k \rangle
+  \frac{\mu}{2} \E \|\vx_n^k-\bar\vx^k\|^2  \nonumber \\
  &+ \E [ f_n(\bar\vx_k) - f_n(\vx^*) ] + \frac{\mu}{2} \E \|\vx_n^k-\vx^*\|^2 \nonumber \\
\geq & - M \E \| \bar\vx^k-\vx^k_n \|   + \E [ f_n(\bar\vx_k) - f_n(\vx^*) ] + \frac{\mu}{4} \E \|\bar\vx^k-\vx^*\|^2 . \nonumber
\end{align}
Combining (\ref{in6-1}) and (\ref{in6-2}), we have
\begin{align}
\label{in6-3}
& - \E \langle \nabla f_n(\vx^{k}_n; \xi_n^{k}),\bar\vx^k-\vx^* \rangle  \\
\leq & 2M \E \| \bar\vx^k-\vx^k_n \|  - \E [ f_n(\bar\vx_k) - f_n(\vx^*) ] - \frac{\mu}{4} \E \|\bar\vx^k-\vx^*\|^2 . \nonumber
\end{align}
{\color{black}Substituting (\ref{in6-0-2}) and (\ref{in6-3}) back into (\ref{in6}), we have
\begin{align}
\label{in6-4}
&\E \| \hat\vx^{k}-\vx^* \|^2  \\
\leq&  (1-\frac{\mu}{4} \alpha^k)\E \| \bar\vx^{k}-\vx^* \|^2 -2\alpha^k\E [ f(\bar\vx^{k})-f(\vx^*) ]  \nonumber\\
&+ (\alpha^k)^2\varphi^2  + \frac{D(\sigma^k)^2}{|\RR|} +  \frac{4 \alpha^k M}{|\RR|}\sum_{n\in\RR}\E \| \vx^{k}_{n}-\bar\vx^{k}\| \nonumber \\
& +\frac{16 \alpha^k \max\{M-\tau,0\}^2}{\mu} \nonumber \\
\leq&  (1-\frac{\mu}{4} \alpha^k)\E \| \bar\vx^{k}-\vx^* \|^2 -2\alpha^k\E [ f(\bar\vx^{k})-f(\vx^*) ]  \nonumber\\
&+ (\alpha^k)^2\varphi^2  + \frac{D(\sigma^k)^2}{|\RR|} +  4 \alpha^k M \E \sqrt{H^k} \nonumber\\
& +\frac{16 \alpha^k \max\{M-\tau,0\}^2}{\mu} . \nonumber
\end{align}
}

Applying (\ref{in6-4}) to (\ref{in5}) yields
\begin{align}
\label{b7}
            &\E \| \bar\vx^{k+1}-\vx^* \|^2  \\
            \leq & \frac{(1-\frac{\mu}{4}\alpha^k)}{a}\E \| \bar\vx^{k}-\vx^* \|^2 -\frac{2\alpha^k}{a}\E [ f(\bar\vx^{k})-f(\vx^*) ]    \nonumber\\
            &+\frac{(\alpha^k)^2\varphi^2}{a}+ \frac{D(\sigma^k)^2}{a|\RR|} +\frac{8\rho^2}{1-a}  \E\max_{n\in  \RR} \|\tilde{\vx}_{n,n}^{k} -  \hat\vx^{k}  \|^2  \nonumber\\
            &+ \frac{2 \chi^2}{(1-a)\vert \RR \vert}\sum_{n\in  \RR} \E \| \tilde{\vx}^{k}_{n,n}-\hat\vx^{k}\|^2 + \frac{4 \alpha^k M \E \sqrt{H^k}}{a}   \nonumber \\
            & {\color{black}+\frac{16 \alpha^k \max\{M-\tau,0\}^2}{\mu a} .} \nonumber
\end{align}

Again, as we do not assume Lipschitz continuous stochastic gradient and bounded stochastic gradient variance, we use Assumption \ref{assumption:gradients} to bound $\E \max_{n\in\RR}  \|\tilde{\vx}^{k}_{n,n}-\hat\vx^{k}\|^2$. For any $v \in (0,1)$, we have an inequality that is similar to (\ref{ineq2}), given by
   {\color{black} \begin{align}
        \label{ineq1}
            & \E \max_{n\in\RR}  \|\tilde{\vx}^{k}_{n,n}-\hat\vx^{k}\|^2         \\
        \le & \frac{1}{1-v}\E\max_{n\in\RR}  \| \vx^{k}_{n}-\bar\vx^{k}\|^2 + \frac{2Dln|\RR|(\sigma^k)^2}{v} \nonumber\\
        & + \frac{(\alpha^k)^2}{v} \E \| \tilde \nabla f_n(\vx^{k}_n; \xi_n^{k}) - \frac{1}{\vert \RR \vert} \sum_{m\in\RR} \tilde\nabla f_m(\vx^{k}_m; \xi_m^{k})  \|^2   \nonumber\\
        \le& \frac{1}{1-v}\E \max_{n\in\RR}  \| \vx^{k}_{n}-\bar\vx^{k}\|^2
        +\frac{4(\alpha^k)^2\varphi^2+2Dln|\RR|(\sigma^k)^2}{v},   \nonumber
    \end{align}}
where the first inequality holds true because
\begin{align}
    \label{b7-1}
    &\E \max_{n \in\RR}\|\ve^{k}_{n} - \frac{1}{\vert \RR \vert} \sum_{m\in\RR} \mathbf{e}_m^{k} \|^2 \\
     =& \E \max_{n \in\RR} \|(1- \frac{1}{\vert \RR \vert})\ve^{k}_{n} - \frac{1}{\vert \RR \vert} \sum_{m\in\RR \setminus \{n\}} \mathbf{e}_m^{k} \|^2  \nonumber \\
    \leq& \E \max_{n \in\RR}\|(1- \frac{1}{\vert \RR \vert})\ve^{k}_{n}\|^2+ \frac{|\RR|-1}{\vert \RR \vert^2} D(\sigma^k)^2  \nonumber\\
    \leq& 2ln|\RR|(1- \frac{1}{\vert \RR \vert})^2D(\sigma^k)^2 + \frac{|\RR|-1}{\vert \RR \vert^2} D(\sigma^k)^2  \nonumber\\
    \leq & 2ln|\RR| D(\sigma^k)^2. \nonumber
\end{align}

Applying (\ref{ineq1}) and (\ref{ineq2}) to (\ref{b7}), we can obtain
\begin{align}
\label{b8}
            &\E \| \bar\vx^{k+1}-\vx^* \|^2 \\
            \leq & \frac{1-\frac{\mu}{4} \alpha^k}{a}\E \| \bar\vx^{k}-\vx^* \|^2 -\frac{2\alpha^k}{a}\E [ f(\bar\vx^{k})-f(\vx^*) ]    \nonumber\\
            &+\frac{(\alpha^k)^2\varphi^2}{a}+ \frac{D(\sigma^k)^2}{a|\RR|} +\frac{16\rho^2}{1-a} \E \max_{n\in\RR}  \| \vx^{k}_{n}-\bar\vx^{k}\|^2  \nonumber\\
            &+ \frac{4 \chi^2}{(1-a)\vert \RR \vert}\sum_{n\in\RR}\E \| \vx^{k}_{n}-\bar\vx^{k}\|^2 + \frac{4 \alpha^k M \E \sqrt{H^k}}{a} \nonumber\\
            &+ \frac{16\rho^2+4 \chi^2}{1-a}  4(\alpha^k)^2\varphi^2
            + \frac{32ln|\RR|\rho^2+4 \chi^2}{1-a}  D(\sigma^k)^2  \nonumber \\
            & {\color{black}+\frac{16 \alpha^k \max\{M-\tau,0\}^2}{\mu a} .} \nonumber
    \end{align}
Rearranging the terms of (\ref{b8}), letting $a=1-\frac{\mu \alpha^k}{8}$ and using the fact that $ 1- \frac{\mu}{4}\alpha^k < (1-\frac{\mu\alpha^k}{8})^2$, we obtain
\begin{align}
\label{b9}
            & \frac{1}{a} \E [ f(\bar\vx^{k})-f(\vx^*) ] \\
            \leq & \frac{1}{2\alpha^k}\lp 1-\frac{\mu \alpha^k}{8}\rp\E \| \bar\vx^{k}-\vx^* \|^2  -\frac{1}{2\alpha^k} \E \| \bar\vx^{k+1}-\vx^* \|^2  \nonumber\\
            &+\frac{\alpha^k \varphi^2}{2a}+ \frac{DC\sigma^k}{2a|\RR|} +\frac{8\rho^2}{(1-a) \alpha^k} \E\max_{n\in\RR}  \| \vx^{k}_{n}-\bar\vx^{k}\|^2  \nonumber\\
            &+ \frac{2  \chi^2}{(1-a) \alpha^k\vert \RR \vert}\sum_{n\in\RR} \E \| \vx^{k}_{n}-\bar\vx^{k}\|^2 + \frac{2 M \E \sqrt{H^k}}{a}  \nonumber\\
            &+\frac{16\rho^2+4\chi^2}{1-a}  2\alpha^k\varphi^2
            + \frac{16ln|\RR|\rho^2+2 \chi^2}{(1-a)\alpha^k}  D(\sigma^k)^2  \nonumber \\
            & {\color{black}+\frac{8 \max\{M-\tau,0\}^2}{\mu a } . }\nonumber
    \end{align}
{\color{black} Because $\max_{n\in\RR} \| \vx^{k}_{n}-\bar\vx^{k}\|^2 \leq \sum_{n\in\RR} \| \vx^{k}_{n}-\bar\vx^{k}\|^2 $ and $\frac{1}{a}=\frac{k+k_0}{k+k_0-1} \in (1,2)$, setting $\alpha^k=\frac{8}{\mu(k+k_0)}$  and $\sigma^k=C\alpha^k$ in (\ref{b9}) yields
\begin{align}
\label{b10}
            \hspace{-1em}& \E [ f(\bar\vx^{k})-f(\vx^*) ]
            \leq  \frac{1}{a} \E [ f(\bar\vx^{k})-f(\vx^*) ]  \\
            \leq& \frac{\mu (k+k_0-1)}{16}\E \| \bar\vx^{k}-\vx^* \|^2  -\frac{\mu (k+k_0)}{16} \E \| \bar\vx^{k+1}-\vx^* \|^2  \nonumber\\
            &+\frac{8 \varphi^2}{\mu(k+k_0)}+ \frac{8 DC^2 }{ \mu |\RR| (k+k_0) } + 4 M \E \sqrt{H^k}     \nonumber\\
            &+\frac{(256\rho^2+64\chi^2)\varphi^2}{\mu}
            + \frac{(128ln|\RR|\rho^2+16 \chi^2)DC^2}{\mu }  \nonumber \\
             & +\frac{\mu(4\rho^2 |\RR |+ \chi^2)(k+k_0)^2}{4}\E H^k  {\color{black}+\frac{16 \max\{M-\tau,0\}^2}{\mu } .} \nonumber
    \end{align}

Summing up (\ref{b10}) from  iteration $0$ to iteration $k$, we have
\begin{align}
\label{b12-0}
           &\sum_{k'=0}^{k} \E [ f(\bar\vx^{k'})-f(\vx^*) ] \\
            \hspace{-2em} \leq&   \frac{\mu  (k_0-1)}{16} \| \bar\vx^{0}-\vx^* \|^2  - \frac{\mu  (k+k_0)}{16} \E \| \bar\vx^{k+1}-\vx^* \|^2 \nonumber\\
            & + \sum_{k'=0}^{k} \frac{8 \varphi^2}{\mu(k'+k_0)}+ \sum_{k'=0}^{k} \frac{8 DC^2 }{ \mu |\RR| (k'+k_0) }  +\sum_{k'=0}^{k} 4M\E \sqrt{H^{k'}}   \nonumber \\
            &+\sum_{k'=0}^{k} \frac{(256\rho^2+64\chi^2)\varphi^2}{\mu}
            + \sum_{k'=0}^{k} \frac{(128ln|\RR|\rho^2+16 \chi^2)DC^2}{\mu }  \nonumber \\
            &+\sum_{k'=0}^{k}  \frac{\mu(4\rho^2 |\RR |+ \chi^2)(k'+k_0)^2}{4}\E H^k  \nonumber \\
            &{\color{black}+\sum_{k'=0}^{k} \frac{8 \max\{M-\tau,0\}^2}{\mu } .} \nonumber
\end{align}

Since $\E [ f(\bar\vx^{k'})-f(\vx^*) ] \geq \E [f_{best}^k-f\lp\vx^*\rp]$, according to (\ref{b12-0}) we have
\begin{align}
\label{b13}
            &\E [f_{best}^k-f\lp\vx^*\rp] \\
            &\leq \frac{\mu (k_0-1)}{16(k+1)} \| \bar\vx^{0}-\vx^* \|^2 \nonumber  + \frac{\sum_{k'=0}^{k} 4M  \E \sqrt{H^{k'}}}{k+1}  \\
            &+\frac{8\varphi^2 ln(k+k_0)}{\mu(k+1)} + \frac{8DC^2 ln(k+k_0)}{\mu |\RR|(k+1)} \nonumber \\
            &+\frac{(256\rho^2+64 \chi^2)\varphi^2}{\mu}+\frac{(128ln|\RR|\rho^2+16 \chi^2)DC^2}{\mu} \nonumber\\
            &+\frac{\sum_{k'=0}^{k} \mu (4\rho^2|\RR|+\chi^2) (k'+k_0)^2  \E H^{k'}}{4(k+1)} \nonumber \\
            & {\color{black}+ \frac{16 \max\{M-\tau,0\}^2}{\mu },} \nonumber
    \end{align}
which completes the proof.}
\end{proof}

\section{Contraction Constants and Virtual Mixing Matrices of Decentralized Robust Aggregation Rules}
\label{dec-proof}

\subsection{Trimmed Mean}
\begin{proof}
We begin by analyzing dimension $d$ of the aggregated result and analyze two cases.

\noindent \textbf{Case 1: Agent $n$ removes all Byzantine messages.}
Suppose that agent $n$ removes all Byzantine messages in dimension $d$. This means $[\mathcal{U}_n]_d \cap \B_n = \varnothing $ and $[\mathcal{U}_n]_d \subset \RR_n \cup \{n\}$, and thus we have
\begin{align}
\hspace{-1em}    & [TM(\tilde{\vx}_{n,n}, \{\tilde{\vx}_{m,n}\}_{m\in \mathcal{R}_n\cup\B_n})]_d  - [\hat \vx_n]_d   \\
\hspace{-1em}    =&  \frac{1}{ \vert \N_n \vert - 2q_n+1 }\sum_{m \in [\mathcal{U}_n]_d} [\tilde{\vx}_{m,n}]_d - \frac{1}{\vert \RR_n\vert+1}\sum_{m \in \RR_n \cup \{n\}} [\tilde{\vx}_{m,n}]_d \nonumber\\
\hspace{-1em}    =&  \frac{1}{ \vert \N_n \vert - 2q_n+1 }\sum_{m \in [\mathcal{U}_n]_d} [\tilde{\vx}_{m,n}]_d \nonumber\\
\hspace{-1em}    &-  \frac{1}{\vert \RR_n\vert+1}\lp \sum_{m \in [\mathcal{U}_n]_d} [\tilde{\vx}_{m,n}]_d + \sum_{m \in \RR_n \cup \{n\} \setminus [\mathcal{U}_n]_d} [\tilde{\vx}_{m,n}]_d \rp. \nonumber
\end{align}
Because $\vert \RR_n\vert+1=\vert \N_n \vert - q_n+1$, it holds that
\begin{align}
    & [TM(\tilde{\vx}_{n,n}, \{\tilde{\vx}_{m,n}\}_{m\in \mathcal{R}_n\cup\B_n})]_d  - [\hat \vx_n]_d   \\
    =&   \frac{q_n}{ (\vert \N_n \vert - 2q_n+1)(\vert \N_n \vert - q_n+1) }\sum_{m \in [\mathcal{U}_n]_d} [\tilde{\vx}_{m,n}]_d  \nonumber \\
     &-  \frac{1}{\vert \N_n \vert - q_n+1}\sum_{m \in \RR_n \cup \{n\} \setminus [\mathcal{U}_n]_d} [\tilde{\vx}_{m,n}]_d.  \nonumber
\end{align}

For convenience, denote
$$\hat\vx_{[\mathcal{U}_n]_d}:=TM(\tilde{\vx}_{n,n}, \{\tilde{\vx}_{m,n}\}_{m\in \mathcal{R}_n\cup\B_n})]_d,$$
$$\hat\vx_{\RR_n \cup \{n\} \setminus [\mathcal{U}_n]_d}:=\frac{1}{q_n} \sum_{m \in \RR_n \cup \{n\} \setminus [\mathcal{U}_n]_d} [\tilde{\vx}_{m,n}]_d.$$
Thus, we have
\begin{align}
    &\Vert [TM(\tilde{\vx}_{n,n}, \{\tilde{\vx}_{m,n}\}_{m\in \mathcal{R}_n\cup\B_n})]_d  - [\hat \vx_n]_d  \Vert \\
    =&\frac{q_n}{\vert \N_n \vert - q_n+1} \lnorm \hat\vx_{[\mathcal{U}_n]_d} -\hat\vx_{\RR_n \cup \{n\} \setminus [\mathcal{U}_n]_d}  \rnorm  \nonumber\\
    \leq &\frac{q_n}{\vert \N_n \vert - q_n+1} \max_{m \in [\mathcal{U}_n]_d} \lnorm [\tilde{\vx}_{m,n}]_d -\hat\vx_{\RR_n \cup \{n\} \setminus [\mathcal{U}_n]_d} \rnorm \nonumber\\
    \leq &\frac{q_n}{\vert \N_n \vert - q_n+1} \max_{m \in \RR_n \cup \{n\}} \lnorm [\tilde{\vx}_{m,n}]_d -\hat\vx_{\RR_n \cup \{n\} \setminus [\mathcal{U}_n]_d}\rnorm \nonumber\\
    \leq &\frac{2q_n}{\vert \N_n \vert - q_n+1}  \max_{m \in \RR_n \cup \{n\}} \lnorm [\tilde{\vx}_{m,n}]_d -[\hat \vx_n]_d\rnorm  \nonumber
\end{align}

\noindent \textbf{Case 2: Agent $n$ cannot remove all Byzantine messages.}
Suppose that agent $n$ cannot remove all Byzantine messages in dimension $d$. It means that $[\mathcal{U}_n]_d \cap \B_n \neq \varnothing $. Denote $ \mathcal{U}_{\B_d} := [\mathcal{U}_n]_d \cap \B_n$ and $ \mathcal{U}_{\RR_d} := [\mathcal{U}_n]_d \cap \RR_n \cup \{n\}$. Also denote
$$\hat\vx_{\mathcal{U}_{\RR_d}} := \frac{1}{\vert \mathcal{U}_{\RR_d} \vert} \sum_{m \in \mathcal{U}_{\RR_d}} [\tilde{\vx}_{m,n}]_d,$$
$$\hat\vx_{\mathcal{U}_{\B_d}} := \frac{1}{\vert \mathcal{U}_{\B_d} \vert} \sum_{m \in \mathcal{U}_{\B_d}} [\tilde{\vx}_{m,n}]_d.$$

With these preparations, we analyze
\begin{align}
\label{e6}
        &\| [TM(\tilde{\vx}_{n,n}, \{\tilde{\vx}_{m,n}\}_{m\in \mathcal{R}_n\cup\B_n})]_d  - [\hat \vx_n]_d \| \\
        =&\| \hat\vx_{[\mathcal{U}_n]_d} - [\hat \vx_n]_d \| \nonumber\\
        \leq & \| \hat\vx_{[\mathcal{U}_n]_d} - \hat\vx_{\mathcal{U}_{\RR_d}} \| + \| \hat\vx_{\mathcal{U}_{\RR_d}} - [\hat \vx_n]_d \|. \nonumber
\end{align}

To bound the first term at the RHS of (\ref{e6}), we recall that
\begin{align}
\label{e4}
        &[TM(\tilde{\vx}_{n,n}, \{\tilde{\vx}_{m,n}\}_{m\in \mathcal{R}_n\cup\B_n})]_d = \hat\vx_{[\mathcal{U}_n]_d} \\
        =& (1-\frac{\vert \mathcal{U}_{\B_d} \vert}{\vert \N_n \vert - 2q_n+1}) \hat\vx_{\mathcal{U}_{\RR_d}} +\frac{\vert \mathcal{U}_{\B_d} \vert}{\vert \N_n \vert - 2q_n+1}\hat\vx_{\mathcal{U}_{\B_d}}. \nonumber
\end{align}
As there  exist certain messages $[\tilde{\vx}_{m,n}]_d$ from non-Byzantine agents that are discarded during the aggregation process, we obtain
\begin{align}
\label{e5}
      \|  \hat\vx_{\mathcal{U}_{\B_d}} - \hat\vx_{\mathcal{U}_{\RR_d}} \| \leq \max_{m \in \RR_n \cup \{n\}} \| [\tilde{\vx}_{m,n}]_d - \hat\vx_{\mathcal{U}_{\RR_d}} \|.
\end{align}
Substituting (\ref{e4}) and (\ref{e5}) into the first term at the RHS of (\ref{e6}) yields
\begin{align}
\label{e7}
       & \| \hat\vx_{[\mathcal{U}_n]_d} - \hat\vx_{\mathcal{U}_{\RR_d}} \| \\
       =& \frac{\vert \mathcal{U}_{\B_d} \vert}{\vert \N_n \vert - 2q_n+1} \| \hat\vx_{\mathcal{U}_{\B_d}} - \hat\vx_{\mathcal{U}_{\RR_d}}  \| \nonumber\\
       \leq &  \frac{\vert \mathcal{U}_{\B_d} \vert}{\vert \N_n \vert - 2q_n+1}   \max_{m \in \RR_n \cup \{n\}} \| [\tilde{\vx}_{m,n}]_d - \hat\vx_{\mathcal{U}_{\RR_d}} \|. \nonumber
\end{align}

For the second term at the RHS of (\ref{e6}), by Lemma 3 in \cite{wu2022byzantine}, we have
\begin{align}
\label{e8}
        &\| \hat\vx_{\mathcal{U}_{\RR_d}} - [\hat \vx_n]_d \|  \\
        \leq &  \frac{2q_n}{\vert \N_n \vert - q_n+1} \max_{m \in \RR_n \cup \{n\} \setminus  \mathcal{U}_{\RR_d}} \| [\tilde{\vx}_{m,n}]_d - \hat\vx_{\mathcal{U}_{\RR_d}}  \| \nonumber \\
         \leq  & \frac{2q_n}{\vert \N_n \vert - q_n+1} \max_{m \in \RR_n \cup \{n\} } \| [\tilde{\vx}_{m,n}]_d - \hat\vx_{\mathcal{U}_{\RR_d}} \|. \nonumber
\end{align}

Substituting (\ref{e7}) and (\ref{e8}) into (\ref{e6}) yields
\begin{align}
        &\| [TM(\tilde{\vx}_{n,n}, \{\tilde{\vx}_{m,n}\}_{m\in \mathcal{R}_n\cup\B_n})]_d  - [\hat \vx_n]_d \| \\
        \leq & \lp \frac{\vert \mathcal{U}_{\B_d} \vert}{\vert \N_n \vert - 2q_n+1} +\frac{2q_n}{\vert \N_n \vert - q_n+1} \rp \nonumber\\
        &\max_{m \in \RR_n \cup \{n\}} \| [\tilde{\vx}_{m,n}]_d - \hat\vx_{\mathcal{U}_{\RR_d}} \| \nonumber\\
        \leq & 2 \lp \frac{\vert \mathcal{U}_{\B_d} \vert}{\vert \N_n \vert - 2q_n+1} +\frac{2q_n}{\vert \N_n \vert - q_n+1} \rp  \nonumber\\
        & \cdot \max_{m \in \RR_n \cup \{n\}} \| [\tilde{\vx}_{m,n}]_d - [\hat\vx_n]_d \|. \nonumber
\end{align}

By combining Case 1 and Case 2, we can deduce that
\begin{align}
\label{e10}
        &\| [TM(\tilde{\vx}_{n,n}, \{\tilde{\vx}_{m,n}\}_{m\in \mathcal{R}_n\cup\B_n})]_d  - [\hat \vx_n]_d \| \\
        \leq & 2 \lp \frac{\vert \mathcal{U}_{\B_d} \vert}{\vert \N_n \vert - 2q_n+1} +\frac{2q_n}{\vert \N_n \vert - q_n+1} \rp \nonumber \\
        &\cdot \max_{m \in \RR_n \cup \{n\}} \| [\tilde{\vx}_{m,n}]_d - [\hat\vx_n]_d \|.  \nonumber
\end{align}

Next, we extend the scalar scenario to the vector scenario. It holds that
\begin{align}
\label{e11}
        & \sum_{d=1}^D \max_{m \in \RR_n \cup \{n\}} \| [\tilde{\vx}_{m,n}]_d - [\hat\vx_n]_d \| ^2 \\
        \leq &  \sum_{d=1}^D \max_{m \in \RR_n \cup \{n\}} \| \tilde\vx_{m,n} - \hat\vx_n \| ^2 \nonumber\\
        \leq & D \max_{m \in \RR_n \cup \{n\}} \| \tilde\vx_{m,n} - \hat\vx_n \| ^2. \nonumber
\end{align}
On the other hand, we also have
\begin{align}
\label{e12}
        & \sum_{d=1}^D \max_{m \in \RR_n \cup \{n\}} \| [\tilde{\vx}_{m,n}]_d - [\hat\vx_n]_d \| ^2 \\
        \leq &  \sum_{d=1}^D \sum_{m \in \RR_n \cup \{n\}} \| [\tilde{\vx}_{m,n}]_d - [\hat\vx_n]_d \| ^2 \nonumber\\
        = & \sum_{m \in \RR_n \cup \{n\}} \sum_{d=1}^D \| [\tilde{\vx}_{m,n}]_d - [\hat\vx_n]_d \| ^2 \nonumber\\
        =  & \sum_{m \in \RR_n \cup \{n\}} \| \tilde\vx_{m,n} - \hat\vx_n \| ^2  \nonumber\\
        \leq & (\vert \RR_n\vert+1) \max_{m \in \RR_n \cup \{n\}} \| \tilde\vx_{m,n} - \hat\vx_n \| ^2.  \nonumber
\end{align}
Combining (\ref{e11}) and (\ref{e12}), we can obtain
\begin{align}
\label{e13}
        & \sum_{d=1}^D \max_{m \in \RR_n \cup \{n\}} \| [\tilde{\vx}_{m,n}]_d - [\hat\vx_n]_d \| ^2 \\
        \leq & \min\{D,(\vert \RR_n \vert+1)\} \max_{m \in \RR_n \cup \{n\}} \| \tilde\vx_{m,n} - \hat\vx_n \| ^2.  \nonumber
\end{align}

Substituting (\ref{e13}) into (\ref{e10}), we have
\begin{align}
\label{e14}
         &\| TM(\tilde{\vx}_{n,n}, \{\tilde{\vx}_{m,n}\}_{m\in \mathcal{R}_n\cup\B_n})  - \hat \vx_n \|^2 \\
       = & \sum_{d=1}^D \| [TM(\tilde{\vx}_{n,n}, \{\tilde{\vx}_{m,n}\}_{m\in \mathcal{R}_n\cup\B_n})]_d  - [\hat \vx_n]_d \|^2  \nonumber\\
        \leq & 4 \lp \frac{\vert \mathcal{U}_{\B_d} \vert}{\vert \N_n \vert - 2q_n+1} +\frac{2q_n}{\vert \N_n \vert - q_n+1} \rp^2  \nonumber\\
        & \cdot \sum_{d=1}^D \max_{m \in \RR_n \cup \{n\}} \| [\tilde{\vx}_{m,n}]_d - [\hat\vx_n]_d \|^2  \nonumber\\
        \leq & 4 \lp \frac{\vert \mathcal{U}_{\B_d} \vert}{\vert \N_n \vert - 2q_n+1} +\frac{2q_n}{\vert \N_n \vert - q_n+1} \rp^2 \nonumber\\
        & \cdot \min\{D,(\vert \RR_n \vert+1)\} \max_{m \in \RR_n \cup \{n\}} \| \tilde\vx_{m,n} - \hat\vx_n \| ^2.  \nonumber
\end{align}
Taking the square roots for both sides of (\ref{e14}) yields
\begin{align}
\label{e15}
         &\| TM(\tilde{\vx}_{n,n}, \{\tilde{\vx}_{m,n}\}_{m\in \mathcal{R}_n\cup\B_n})  - \hat \vx_n \| \\
        \leq & 2 \lp \frac{\vert \mathcal{U}_{\B_d} \vert}{\vert \N_n \vert - 2q_n+1} +\frac{2q_n}{\vert \N_n \vert - q_n+1} \rp \nonumber\\
        & \cdot \min\{D^{\frac{1}{2}},(\vert \RR_n \vert+1) ^{\frac{1}{2}}\} \max_{m \in \RR_n \cup \{n\}} \| \tilde\vx_{m,n} - \hat\vx_n \|  \nonumber \\
         \leq & 2 \lp \frac{q_n}{\vert \N_n \vert - 2q_n+1} +\frac{2q_n}{\vert \N_n \vert - q_n+1} \rp\nonumber\\
        & \cdot \min\{D^{\frac{1}{2}},(\vert \RR_n \vert+1) ^{\frac{1}{2}}\} \max_{m \in \RR_n \cup \{n\}} \| \tilde\vx_{m,n} - \hat\vx_n \|,  \nonumber
\end{align}
which completes the proof.
\end{proof}

\subsection{SCC}
\begin{proof}
For notational convenience, define $\tilde z_{m,n} = \tilde{\vx}_{n,n} + Clip(\tilde{\vx}_{m,n}-\tilde{\vx}_{n,n},\tau_n)$. First, we analyze

    \begin{align}
            &\Vert SCC(\tilde{\vx}_{n,n}, \{\tilde{\vx}_{m,n}\}_{m\in \mathcal{R}_n\cup\B_n}) - \hat \vx_n  \Vert ^2 \\
             = & \lnorm \sum_{m \in \mathcal{N}_n\cup \{n\}} w'_{nm} {\tilde z_{m,n}}  - \sum_{m \in \mathcal{R}_n\cup \{n\}} w_{nm} \tilde\vx_{m,n}  \rnorm ^2. \nonumber
    \end{align}

Considering the virtual mixing matrix defined in Lemma \ref{scc-rho}, we know
\begin{align}
\label{d5}
         &\Vert SCC(\tilde{\vx}_{n,n}, \{\tilde{\vx}_{m,n}\}_{m\in \mathcal{R}_n\cup\B_n}) - \hat \vx_n  \Vert ^2 \\
          =  & \lnorm \sum_{m \in \mathcal{N}_n\cup \{n\}} w'_{nm} {\tilde z_{m,n}}  - \sum_{m \in \mathcal{R}_n\cup \{n\}} w_{nm} \tilde\vx_{m,n}  \rnorm ^2  \nonumber\\
          = & \lnorm \sum_{m \in \mathcal{R}_n} w'_{nm} ({\tilde z_{m,n}}  -  \tilde\vx_{m,n} )
           +   \sum_{m \in \mathcal{B}_n} w'_{nm} ({\tilde z_{m,n}}  -  \tilde \vx_{n,n} ) \rnorm ^2 \nonumber\\
           \leq & 2\lnorm \sum_{m \in \mathcal{R}_n} w'_{nm} ({\tilde z_{m,n}}  -  \tilde\vx_{m,n}  ) \rnorm ^2
           +  2\lnorm \sum_{m \in \mathcal{B}_n} w'_{nm} ({\tilde z_{m,n}}  - \tilde \vx_{n,n} ) \rnorm ^2.  \nonumber
    \end{align}

For the first term at the RHS of (\ref{d5}), we have
\begin{align}
\label{d6}
         & 2\lnorm \sum_{m \in \mathcal{R}_n} w'_{nm} ({\tilde z_{m,n}}  -  \tilde\vx_{m,n}  ) \rnorm ^2  \\
         \leq & 2\lp \sum_{m \in \mathcal{R}_n} w'_{nm} \lnorm {\tilde z_{m,n}}  -  \tilde\vx_{m,n}  \rnorm  \rp^2 \nonumber\\
         \leq & 2\lp \frac{1}{\tau_n}\sum_{m \in \mathcal{R}_n} w'_{nm} \lnorm  \tilde \vx_{n,n}  -  \tilde\vx_{m,n}  \rnorm ^2 \rp^2. \nonumber
    \end{align}
The first inequality holds true because $\langle \bm{a},\bm{b}\rangle \leq \|\bm{a}\|\|\bm{b}\|$. The second inequality holds true if no clipping is performed, because ${\tilde z_{m,n}}  -  \tilde\vx_{m,n}  = 0$. If clipping is performed, then $\| \tilde \vx_{n,n}  -  \tilde\vx_{m,n}  \| = \tau_n + \| {\tilde z_{m,n}}  -  \tilde\vx_{m,n}  \|$ and the second inequality also holds true since $\| \tilde \vx_{n,n}  -  \tilde\vx_{m,n}  \|-\tau_n \leq \frac{1}{\tau_n} \|  \tilde \vx_{n,n}  -  \tilde\vx_{m,n}  \|^2$ according to $a-b \leq \frac{a^2}{b}$.

For the second term at the RHS of (\ref{d5}), we have
\begin{align}
\label{d7}
         &  2\lnorm \sum_{m \in \mathcal{B}_n} w'_{nm} ({\tilde z_{m,n}}  - \tilde \vx_{n,n} ) \rnorm ^2  \\
         \leq & 2\lp \sum_{m \in \mathcal{B}_n} w'_{nm} \lnorm {\tilde z_{m,n}}  -  \tilde\vx_{m,n}  \rnorm  \rp^2 \nonumber\\
         \leq & 2\lp \sum_{m \in \mathcal{B}_n} w'_{nm}  \tau_n  \rp^2. \nonumber
    \end{align}

Recall that $\tau_n$ is chosen as
\begin{align}
\label{d8}
        \hspace{-1em} \tau_n = \lp \frac{1}{\sum_{m \in \mathcal{B}_n} w'_{nm}} \sum_{m \in \mathcal{R}_n} w'_{nm} \lnorm  \tilde \vx_{n,n}  -  \tilde\vx_{m,n}  \rnorm ^2 \rp^{\frac{1}{2}}.
    \end{align}

Substituting (\ref{d6})--(\ref{d8}) into (\ref{d5}) yields
\begin{align}
\label{d9}
\hspace{-1em}         &\Vert SCC(\tilde{\vx}_{n,n}, \{\tilde{\vx}_{m,n}\}_{m\in \mathcal{R}_n\cup\B_n}) - \hat \vx_n  \Vert ^2  \\
\hspace{-1em}           \leq & 2\lnorm \sum_{m \in \mathcal{R}_n} w'_{nm} ({\tilde z_{m,n}}  -  \tilde\vx_{m,n}  ) \rnorm ^2
           +  2\lnorm \sum_{m \in \mathcal{B}_n} w'_{nm} ({\tilde z_{m,n}}  - \tilde \vx_{n,n} ) \rnorm ^2 \nonumber\\
\hspace{-1em}           \leq & 2\lp\frac{1}{\tau_n} \sum_{m \in \mathcal{R}_n} w'_{nm} \lnorm  \tilde \vx_{n,n}  -  \tilde\vx_{m,n}  \rnorm^2  \rp^2 + 2\lp \sum_{m \in \mathcal{B}_n} w'_{nm}  \tau_n  \rp^2 \nonumber\\
\hspace{-1em}           \leq & 4 \sum_{m \in \mathcal{B}_n}w'_{nm} \sum_{m \in \mathcal{R}_n} w'_{nm} \lnorm  \tilde \vx_{n,n}  -  \tilde\vx_{m,n}  \rnorm^2.  \nonumber
    \end{align}
Note that
\begin{align}
\label{d10}
         & \lnorm  \tilde \vx_{n,n}  -  \tilde\vx_{m,n}  \rnorm^2 \\
         \leq & 2\lnorm  \tilde \vx_{n,n}  -  \hat \vx_n  \rnorm^2 +  2\lnorm  \tilde\vx_{m,n}  -  \hat \vx_n  \rnorm^2 \nonumber\\
         \leq&  4 \max_{m \in \RR_n\cup \{n\}} \lnorm  \tilde\vx_{m,n}  -  \hat \vx_n  \rnorm^2.  \nonumber
    \end{align}
Further substituting (\ref{d10}) into (\ref{d9}) yields
\begin{align}
\label{d11}
         &\Vert SCC(\tilde{\vx}_{n,n}, \{\tilde{\vx}_{m,n}\}_{m\in \mathcal{R}_n\cup\B_n}) - \hat \vx_n  \Vert ^2 \\
          & \leq  16 \sum_{m \in \mathcal{B}_n}w'_{nm} \sum_{m \in \mathcal{R}_n} w'_{nm} \max_{m \in \RR_n\cup \{n\}} \lnorm  \tilde\vx_{m,n}  -  \hat \vx_n  \rnorm^2. \nonumber
    \end{align}
Taking the square roots for both sides of (\ref{d11}) yields
\begin{align}
\label{d12}
         &\Vert SCC(\tilde{\vx}_{n,n}, \{\tilde{\vx}_{m,n}\}_{m\in \mathcal{R}_n\cup\B_n}) - \hat \vx_n  \Vert  \\
          & \leq  4 \sqrt{ \sum_{m \in \mathcal{B}_n}w'_{nm} \sum_{m \in \mathcal{R}_n} w'_{nm}} \max_{m \in \RR_n\cup \{n\}} \lnorm  \tilde\vx_{m,n}  -  \hat \vx_n  \rnorm^2, \nonumber
    \end{align}
which completes the proof.
\end{proof}

\subsection{IOS}
\begin{proof}
When not all Byzantine messages are successfully removed, according to Theorem 3 in \cite{wu2022byzantine} we obtain (\ref{equality:rho-of-ios-2}). On the other hand, when all Byzantine messages are successfully removed, the output of IOS at non-Byzantine agent $n$ is
\begin{align}
    \label{ios-0}
    & IOS(\tilde{\vx}_{n,n}, \{\tilde{\vx}_{m,n}\}_{m\in \mathcal{R}_n\cup\B_n}) \\
    = & \frac{\sum_{m \in \RR_n \cup \{n\}}w'_{nm} \tilde{\vx}_{m,n}}{\sum_{m \in \RR_n \cup \{n\}}w'_{nm}}. \nonumber
\end{align}
When the  $(n,m)$-th entry of the associated virtual mixing matrix $W$ is given by
\begin{align}
    w_{nm} = \begin{cases}
        w_{nn}'+\sum_{b\in\B_n} w_{nb}', &m=n,\\
        w_{nm}', &m\neq n,\\
    \end{cases}
\end{align}
we can rewrite the weighted average $\hat\vx_n$ as
\begin{align}
    \label{ios-1}
    \hat\vx_n =  \sum_{m \in \RR_n \cup \{n\}}w'_{nm} \tilde{\vx}_{m,n}  +   \sum_{b \in \B_n}w'_{nb} \tilde{\vx}_{n,n}.
\end{align}

Thus, $IOS(\tilde{\vx}_{n,n}, \{\tilde{\vx}_{m,n}\}_{m\in \mathcal{R}_n\cup\B_n})$ is a weighted mean over the set $\{\tilde{\vx}_{n,n}, \{\tilde{\vx}_{m,n}\}_{m\in \mathcal{R}_n\cup\B_n}\}$ and $\hat\vx_n$ is a weighted mean over the set $\{\tilde{\vx}_{n,n}, \tilde{\vx}_{n',n'},\{\tilde{\vx}_{m,n}\}_{m\in \mathcal{R}_n\cup\B_n}\}$, where $\tilde{\vx}_{n',n'}=\tilde{\vx}_{n,n}$ but the associated weight is $\sum_{b\in\B_n} w_{nb}'$. Utilizing Lemma 3 in \cite{wu2022byzantine}, we have
\begin{align}
\label{ios-2}
        & \|IOS(\tilde{\vx}_{n,n}, \{\tilde{\vx}_{m,n}\}_{m\in \mathcal{R}_n\cup\B_n})-\hat\vx_n \| \\
        \leq & \frac{\sum_{m \in \B_n}w'_{nm}}{\sum_{m \in \RR_n \cup \{n\}}w'_{nm}} \max_{m \in \RR_n \cup \{n\}} \| \tilde\vx_{m,n} - \hat\vx_n \| \nonumber\\
        \leq & \frac{\sum_{m \in \B_n}w'_{nm}}{1-\sum_{m \in \B_n}w'_{nm}} \max_{m \in \RR_n \cup \{n\}} \| \tilde\vx_{m,n} - \hat\vx_n \| \nonumber\\
        \leq & \frac{\sum_{m \in \cH_n}w'_{nm}}{1-\sum_{m \in \cH_n}w'_{nm}} \max_{m \in \RR_n \cup \{n\}} \| \tilde\vx_{m,n} - \hat\vx_n \|, \nonumber
\end{align}
which completes the proof.
\end{proof}

\section{Additional Numerical Experiments on CIFAR10}
\label{extra-ex}
In addition to the convex logistic regression problem on MNIST, we also test a non-convex ResNet18 training problem on CIFAR10. The CIFAR10 dataset consists of 60,000 color images in 10 classes, with 50,000 training images and 10,000 test images. In the i.i.d. case, these training images are randomly and evenly allocated to all the agents. In the non-i.i.d. case, one class of training images are shared by 4 non-Byzantine agents and each non-Byzantine agent has 4 classes of training images.
We use the same network topology as in the logistic regression task. Hence, all robust aggregation rules have the same $\rho$ and $\chi^2$ as in Table \ref{table:rho-case1}. 
We set the clipping threshold $\tau = 4$.
Furthermore, the theoretical step size $\alpha^{k} = O(\frac{1}{k})$ decreases too rapidly, and thus we use a diminishing step size that decreases by a factor of 0.9 at every 30 iterations. We initialize ResNet18 with a pre-trained model, which is different to the logistic regression task. As a result, the disagreement measure of the non-Byzantine agents starts from zero in the following numerical experiments.

\subsection{Tradeoffs in Different Robust Aggregation Rules}
\begin{figure*}
\centering
\includegraphics[width=7in]{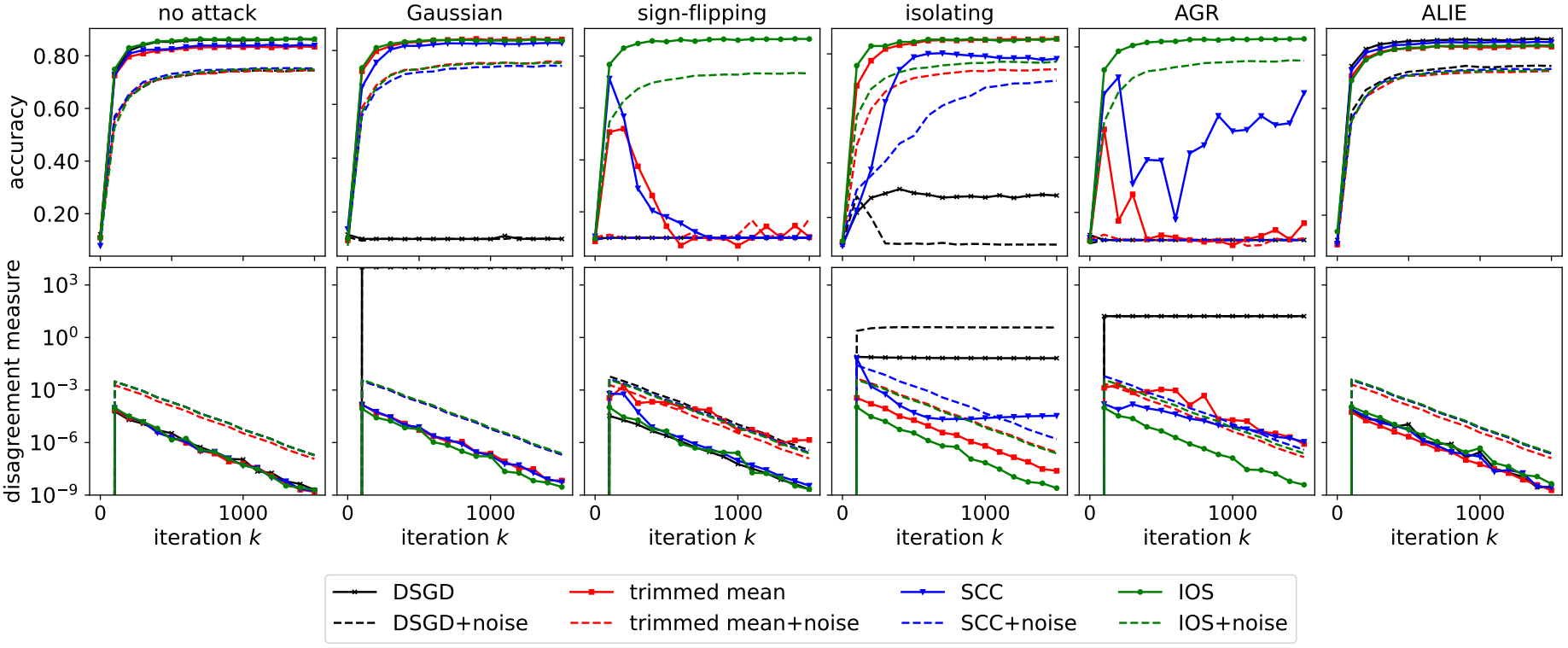}
\caption{Different robust aggregation rules in non-i.i.d. setting on CIFAR10.}\label{cifar10-noniid}
\end{figure*}

We set a fixed noise level $C=0.02$ with total privacy budgets $(\epsilon,\delta)=(68.78,10^{-4})$  and demonstrate the performance of different robust aggregation rules in 
Fig. \ref{cifar10-noniid}.
The observations  align with those on the MNIST dataset, indicating the applicability of our theoretical findings in the non-convex problems.
Notably, when combined with the Gaussian privacy mechanism, all robust aggregation rules experience greater performance degradation due to the larger model dimension, which leads to larger learning errors. This phenomenon matches our analysis.


\subsection{Tradeoffs in Different Noise Levels}
\begin{figure*}
\centering
{\includegraphics[width=7in]{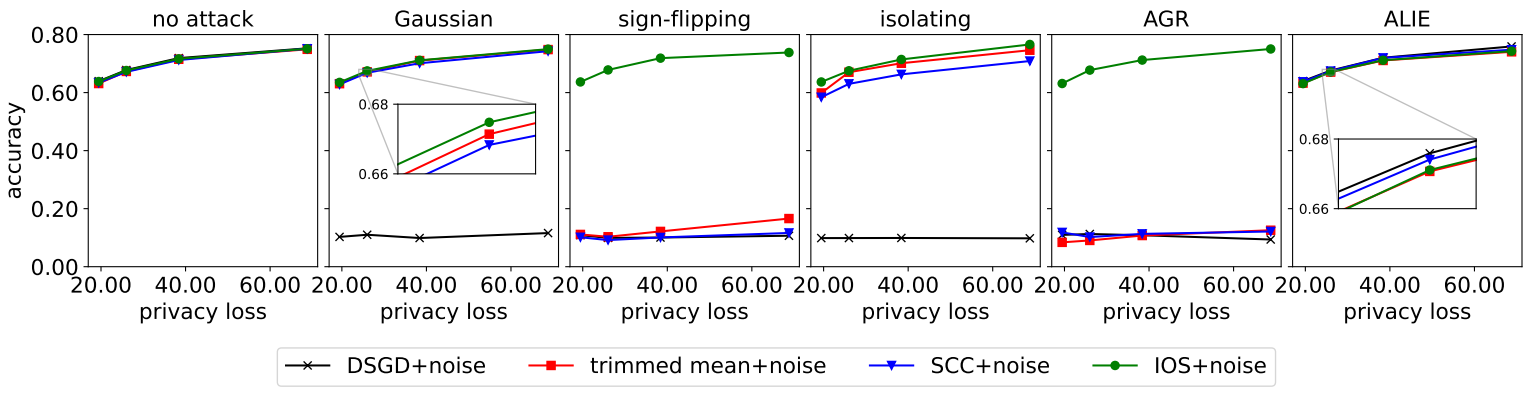}}
\caption{Different robust aggregation rules under different noise levels in non-i.i.d. setting on CIFAR10.}\label{compare-cifar10-noniid}
\end{figure*}

To illustrate the tradeoff between privacy preservation and Byzantine-robustness, we implement different noise levels for different robust aggregation rules in 
Fig. \ref{compare-cifar10-noniid}.  We set the noise levels to $C = 0.02$, $0.03$, $0.04$, $0.05$   and the total privacy budgets after $K=1500$ iterations are $(\epsilon,\delta)=(68.78,10^{-4})$, $(\epsilon,\delta)=(38.39,10^{-4})$, $(\epsilon,\delta)=(25.99,10^{-4})$, $(\epsilon,\delta)=(19.45,10^{-4})$, respectively. Our observations are consistent with those on MNIST.


\subsection{Performance under DLG Attacks}

\begin{figure*}
\centering
{\includegraphics[width=3.5in]{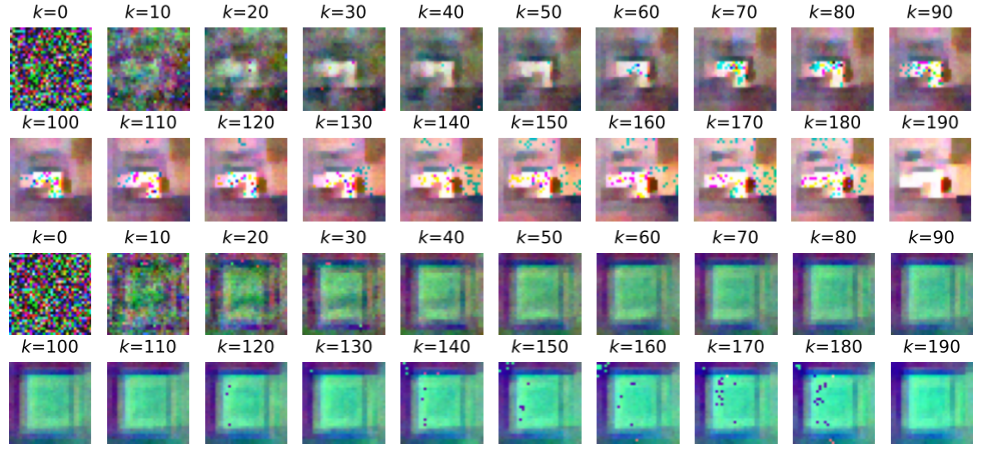}}
\caption{DLG recovery for Byzantine-robust decentralized SGD (TOP) and privacy-preserving and Byzantine-robust decentralized SGD (BOTTOM).}\label{dlg1}
\end{figure*}

To depict the privacy preservation ability of the proposed method, we conduct the \textbf{DLG attacks}. Following the experimental setting on MNIST, we consider a worst-case scenario.
Let an agent be surrounded by curious agents such that its received and transmitted messages are all exposed -- in this case, for any robust aggregation rule the curious agents can always know the output.
Specifically, we let the chosen agent compute its stochastic gradients from the same sample of the image labeled as  ``truck". The goal of the curious agents is to recover this sample. Fig. \ref{dlg1} illustrates that the curious agents can recover the sample for Byzantine-robust decentralized SGD to a certain extent, but fail when the Gaussian noise mechanism is incorporated. It is important to highlight that the high dimensionality of the ResNet18 model renders the DLG attacks less effective. Even a smaller noise level proves to be sufficient for ensuring privacy.



\end{document}